\documentclass[runningheads]{llncs}
\usepackage{graphicx}
\usepackage{tikz}
\usepackage{comment}
\usepackage{amsmath,amssymb} 
\usepackage{color}
\usepackage{subfigure}
\usepackage{graphicx}

\begin{document}
\pagestyle{headings}
\mainmatter
\def\ECCVSubNumber{6968}  

\title{Perception Over Time: Temporal Dynamics for Robust Image Understanding}

\titlerunning{Perception Over Time}
\author{Maryam Daniali 
\and
Edward Kim
}

\authorrunning{M. Daniali, E. Kim.}
\institute{Drexel University, Philadelphia, PA 19104, USA \\
\email{\{md3464,ek826\}@drexel.edu}}

\maketitle

\begin{abstract}
While deep learning surpasses human-level performance in narrow and specific vision tasks, it is fragile and over-confident in classification.  For example, minor transformations in perspective, illumination, or object deformation in the image space can result in drastically different labeling, which is especially transparent via adversarial perturbations.
On the other hand, human visual perception is orders of magnitude more robust to changes in the input stimulus. But unfortunately, we are far from fully understanding and integrating the underlying mechanisms that result in such robust perception.

In this work, we introduce a novel method of incorporating temporal dynamics into static image understanding. We describe a neuro-inspired method that decomposes a single image into a series of coarse-to-fine images that simulates how biological vision integrates information over time. Next, we demonstrate how our novel visual perception framework can utilize this information ``over time'' using a biologically plausible algorithm with recurrent units, and as a result, significantly improving its accuracy and robustness over standard CNNs. We also compare our proposed approach with state-of-the-art models and explicitly quantify our adversarial robustness properties through multiple ablation studies. Our quantitative and qualitative results convincingly demonstrate exciting and transformative improvements over the standard computer vision and deep learning architectures used today.

\end{abstract}

\section{Introduction}\label{sec:introduction}

Human visual perception is remarkably slow. The moment an input stimulus is presented to the time of recognition takes several hundred milliseconds. Neuroscience experiments and recordings offer an explanation; the visual system is performing \textit{recognition over time}, even for static and simple visual scenes. Perceptual clarity increases as bottom-up signals and top-down feedback mechanisms compete and converge to a confident agreement.  In fact, it is precisely this ``slowness'' that makes human perception so accurate and robust.

In contrast, standard deep learning classifiers typically only implement a single feed-forward pass through the network and can be optimized in hardware to be orders of magnitude faster than human recognition. But this speed comes at a cost, i.e., there is no top-down feedback loop nor notion of predictive coding feedback (expectation guiding perception). Thus, the idea that a deep learning network would perceive \textit{static images and scenes over time has never been investigated}, nor have there ever been any studies on how deep learning models might implement temporal dynamics for image classification.

We believe a more biologically inspired model of ``slowness'' will ultimately provide mechanisms and solutions that are robust in general classification, and especially effective against  adversarial examples. 
Thus, in this work, we present a novel architecture that utilizes the idea of classification over time for the robust classification of static images.  In essence, the model ``sees'' a gradual progression of more of the input signal over a generated time series of increasing perceptual clarity extracted from a single input stimulus. The final classification is the culmination of all the information integrated over time. Our contributions are as follows:
\begin{itemize}
    \item [-]We propose a bio-inspired recurrent model that generates a series of images that capture the gradual progression of static image resolution.
    \item [-]We demonstrate a perception framework that can integrate visual information over time to perform more accurate and robust image classification.
    \item [-]To the best of our knowledge, this is the first study that considers temporal dynamics in the visual perception of static images.
    \item [-]Our visual perception framework achieves superior performance compared with the state-of-the-art methods and is more robust to perturbations and adversarial attacks.
\end{itemize}

\section{Related Work}\label{sec:relatedwork}

Our perception of a visual scene changes rapidly over time, even if the scene remains unchanged \cite{hegde2008time}. Although we are far from fully understanding the changes in human visual perception over time, some studies provide considerable evidence on the existence of temporal dynamics in visual recognition \cite{kersten2004object,ma200616neural}. 
Psychophysical studies show that around 150 ms after the stimulus onset, humans acquire the ``gist'' of complex visual scenes, even when the stimulus is presented very briefly. They require longer processing to identify individual objects, and it may even take longer for a more comprehensive semantic understanding of the scene to be encoded into short-term memory \cite{hegde2008time}. Consistent with the timing of perceptual understanding, many studies have suggested that the visual system integrates visual input in a coarse-to-fine (CtF) manner \cite{bar2006top}. The CtF hypothesis states that low-frequency information is processed quickly first, which then projects to high-level visual areas.  Critically, the high-level areas \textit{generate a feedback signal} that guides the processing of the high-frequency input \cite{petras2019coarse}.  David Marr’s work on a functional model of the visual system \cite{marr1979computational} also emphasizes several levels of computation, e.g., primal sketch to 2.5D to 3D representation, mimicked by the cortical areas of the primate visual system.

Processing visual input in a CtF manner helps humans achieve highly robust and accurate perception. Indeed, our own experiences demonstrate that small changes in the input do not change our understanding dramatically. After some amount of information, there is a certain point in time when our brains can detect and identify objects at a very high level of certainty, but prior to this ``aha'' moment that occurs hundreds of milliseconds after stimulus onset, we are (justifiably) neither confident nor certain in object recognition.

At the other end of the recognition spectrum, deep learning is very sensitive and fragile to small changes in the input stimuli and overly confident in classification. Our work described in this manuscript will elucidate these points further.  Additional literature attempts to protect deep learning models from transformations, perturbations, and adversarial attacks, including but not limited to augmenting training data \cite{madry2017towards}, adding stochasticity to the hidden layers \cite{das2017keeping}, and applying preprocessing techniques \cite{xu2017feature,kim2019neuromorphic,guo2017countering}. 
While such techniques are capable of improving the image classification model on a specific task and data \cite{shankar2021image,vuyyuru2020biologically}, even at the cost of heavy computations, research has shown every defense against adversarial attacks has eventually been found to be vulnerable, and there still is a monumental gap between human perception and the current state of machine vision.

\section{Methods}\label{sec:method}


In our work, we take a neuro-inspired approach for robust vision by simulating a series of reflections from an input image, similar to the perceptual process of the human brain. The first step in doing so is to simulate the coarse-to-fine structure of a visual scene by generating components that represent the changes of visual perception over time. Furthermore, by generating these components, we can investigate the robustness of available approaches and design a model inspired by the psychophysical findings on dynamic components in perception over time.

\subsection{Coarse-to-Fine (CtF) Decomposition Methods}\label{subsec:decomposition_methods}
Image decomposition is the general process of separating an input stimulus into a combination of the generators (or causes) of the data.   Decomposition methods have been used in various computer vision applications such as object detection, background subtraction \cite{javed2014robust}, and moving object detection \cite{shakeri2019moving}.  They also have applications in image smoothing and deblurring \cite{firsov2006domain,xu2014domain}. 
In this paper, we introduce a sparse coding model that can faithfully mimic CtF decomposition over time.  We also describe two other approximate decomposition methods of minimal biological fidelity, but more readily available to the general public, e.g. JPEG and Gaussian decomposition.  It is worth mentioning that while studies such as \cite{kim2020modeling} show biological models such as sparse coding are more robust to perturbations than JPEG Compression \cite{das2017keeping} and Gaussian Smoothing \cite{xu2017feature}, based on the available literature, we believe these methods play an important role as baselines.
\vspace{-0.4cm}
\subsubsection{Sparse Coding.}
Sparse coding provides a class of algorithms for finding sparse representations of stimuli, input data. Given only unlabeled data, sparse coding looks for generating a minimal set of components that can reconstruct each input signal as accurately as possible.
In 1997, Olshausen and Field \cite{olshausen1997sparse} introduced sparse coding to explain the sparse and recurrent neural representations in the primary visual cortex. In sparse coding reconstruction, the goal is to minimize the input signal reconstruction error and increase the sparsity by maximizing zero coefficients. These characteristics make sparse coding have a high representative capacity which surpasses the capabilities of dense networks on pairing inputs and outputs. Furthermore, based on its unsupervised nature, it can leverage the availability of unlabeled data. Unlike some other unsupervised learning techniques such as Principal Component Analysis (PCA) \cite{wold1987principal}, Independent Component Analysis (ICA), and Auto-encoders \cite{rumelhart1985learning}, sparse coding can be applied to learning overcomplete basis sets, in which the number of bases is greater than the input dimension \cite{lee2007efficient}.

Sparse coding can be defined using the following objective function, where $x^{(n)}$ represents the input signal, and $a^{(n)}$ is the sparse representation, also known as activation, that can reconstruct the input $x^{(n)}$. 
\begin{equation}
	\min_\Phi \sum^{\mathcal{N}}_{n=1} \min_{a^{(n)}} \frac{1}{2} \| x^{(n)} - \Phi a^{(n)}\|^2_2 + \lambda \|a^{(n)}\|_1
	\label{eq:sparse_coding}
\end{equation}
Here, $\Phi$ is an overcomplete dictionary containing all components that share features to reconstruct the input, and $\hat{x}^{(n)} =\Phi a^{(n)}$ is the reconstructed form. $\lambda$ balances the sparsity versus the reconstruction quality. $n$ is a training element, and there are a total of $\mathcal{N}$ training elements. 

There are different solvers for Equation \ref{eq:sparse_coding}, and, among them, there are some systems of nonlinear differential equations, including but not limited to Iterative Shrinkage and Thresholding Algorithm (ISTA)\cite{daubechies2004iterative} and Locally Competitive Algorithm (LCA)\cite{rozell2007locally}. 
Here, we select the Locally Competitive Algorithm, a bio-inspired technique that evolves the dynamical variables, the membrane potential of the neuron, when an input signal is presented. In this model, the activations of neurons compete and inhibit other units from firing. The neuron's excitatory potential is proportional to the match between the input signal and the dictionary element of that neuron. On the other hand, the inhibitory strength is proportional to the similarity of elements/convolutional patches between the current neuron and other competing neurons, forcing it to decorrelate.

In the LCA algorithm, the active coefficients for a neuron, $m$, with the membrane potential, i.e., the internal state, $u^m$, can be defined as:
\begin{equation}
    a^m = T_\lambda(u^m)=H(u^m-\lambda)u^m
\end{equation}
where $T$ is a soft-threshold transfer function with threshold parameter, $\lambda$, and $H$ is the Heaviside, step, function \cite{abramowitz1964handbook}.  

The ordinary differential equation below determines the dynamics of a neuron, $m$, with an input signal, $I$.
\begin{equation}
 \dot{u}^m =  \frac{1}{\tau} \bigg[ -u^m + (\Phi^T I) - (\Phi^T \Phi a - a^m)\bigg]
 \label{eq:orig}
\end{equation}

 where $\tau$ is the time constant, $-u^m$ is the internal state leakage term, $(\Phi^T I)$ is the driver that charges up the state by the match between the dictionary element and the input signal, here calculated by the inner product between them. $(\Phi^T \Phi a - a^m)$ shows the competition between the set of active neurons proportional to the inner product between dictionary elements, which applies as a lateral inhibition signal. $- a^m$ excludes self-interactions, including self-inhibition. 
 
 In short, using LCA, neurons that are selective to the input stimulus charge up faster, then pass a threshold of activation.  Once they pass the threshold, they begin to compete with other neurons to claim the representation.  Thus sparse coding with the LCA solver creates a sparse representation of selective neurons that compete with each other to represent stimuli \cite{paiton2020selectivity,kim2019Neuromo}. 
\vspace{-0.4cm} 
\subsubsection{Recurrent Sparse Coding Decomposition (RSCD).}
\begin{figure}[!t] 
 \centering
    \subfigure[RSCD model schematic]{\includegraphics[width=4.5cm]{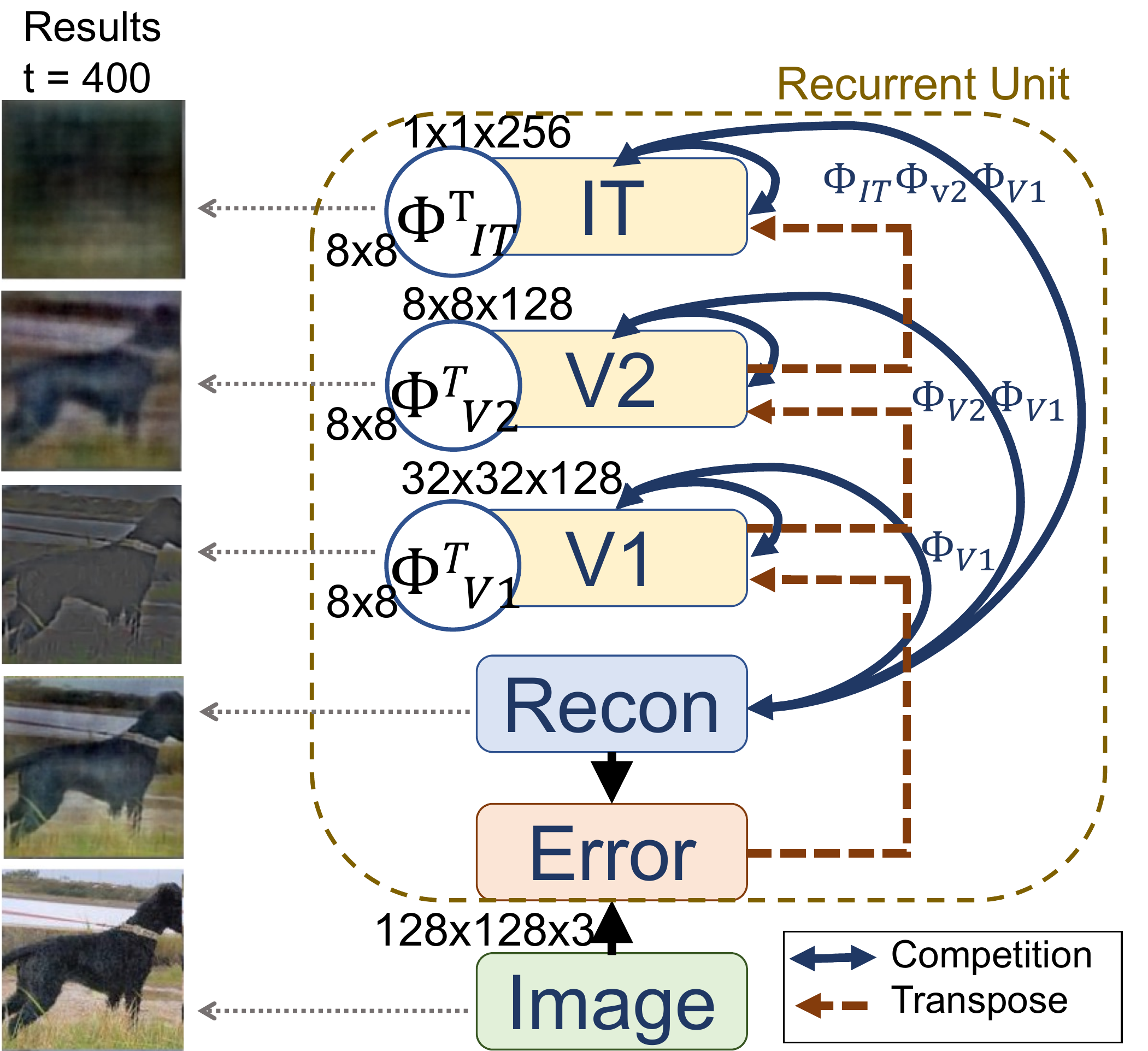}}
    \subfigure[Sample decomposed images]{\includegraphics[width=7.5cm]{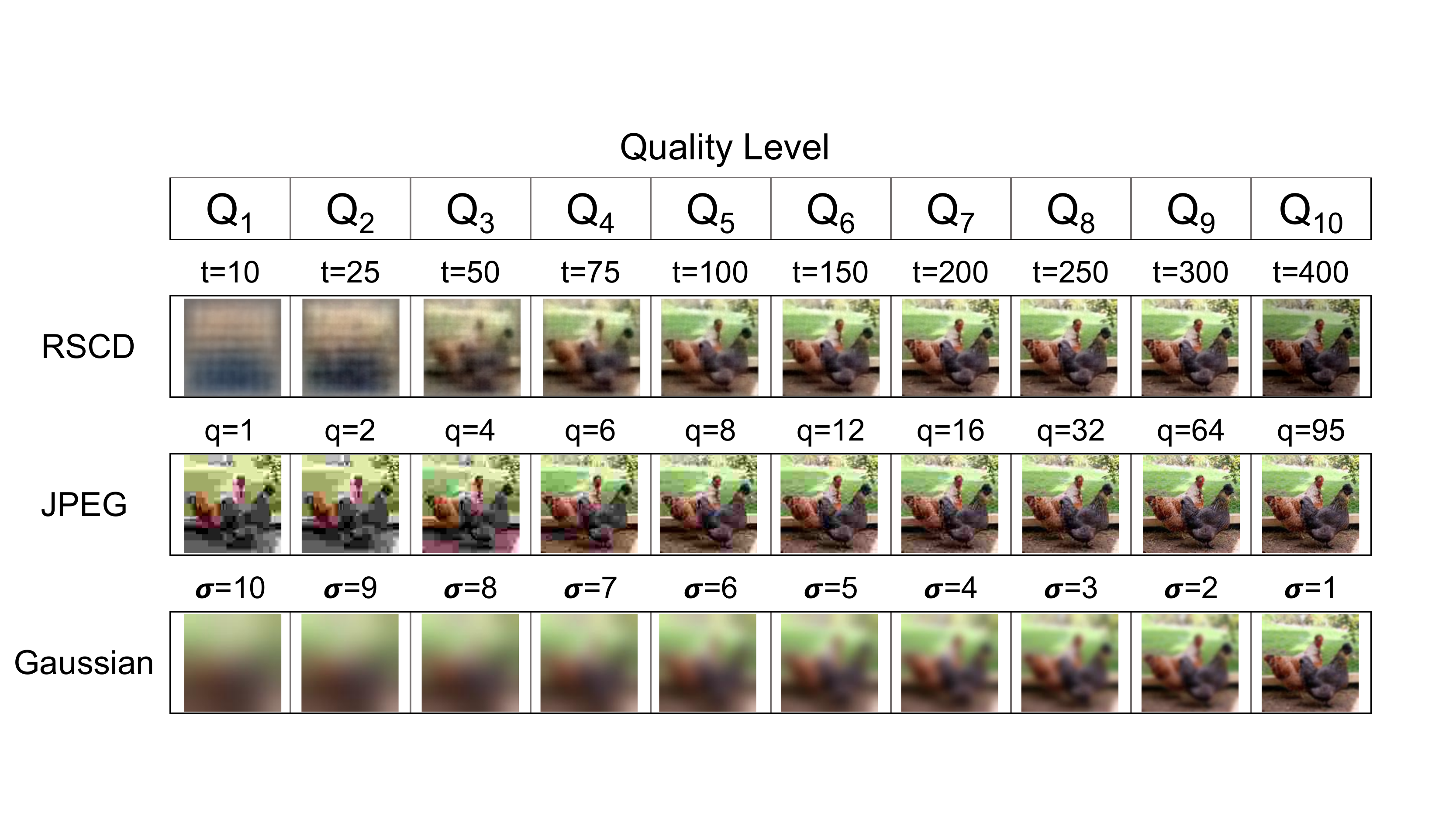}}
  \caption{a) Schematic of our sparse model (RSCD) for image decomposition over time. b) Sample decomposed images with different qualities using RSCD (1st row), JPEG compression (2nd row), and Gaussian smoothing (3rd row).}
  \label{fig:model_sample}
  \end{figure}

We developed a biologically inspired recurrent model that uses selectivity through competition, holistic processing, and top-down feedback to generate image decomposition over time using sparse coding. We call our model Recurrent Sparse Coding Decomposition (RSCD) and use it to decompose images 
in a CtF manner over $t=400$ time steps. Our model schematic is presented in Figure \ref{fig:model_sample}(a), and Equation \ref{eq:sparsecode_RSCD},
\begin{equation}
	\hat{x}^{(n)} = \sum_{k=1}^{K}(\prod_{l=1}^{k} \Phi_{l}) a_{k}^{(n)},
	\label{eq:sparsecode_RSCD}
\end{equation}
where $\hat{x}^{(n)}$ represents the final reconstruction/decomposition of input ${x}^{(n)}$, which we can substitute into Equation \ref{eq:sparse_coding} and dynamically solve using Equation \ref{eq:orig}.  $K$ is the number of layers in the sparse model. For RSCD, we have three layers, and $ k \in \{V1, V2, IT\}$ which roughly match layers involved in the ventral pathway of the human visual cortex used for form recognition and object representation. 

For evaluation purposes, we selected a subset of 10 decomposed images generated by RSCD, namely $ t \in \{10,25,50,75,199,150,200,250,300,400\}$, where $t$ is the timestep of the model.
Figure \ref{fig:model_sample}(b) shows the selected quality levels of one image using the RSCD decomposition method.

\subsection{Approximate CtF decomposition}
While approximation methods do not decompose input stimuli over time and are not biologically plausible, they provide a good approximation and are generally fast. Among them, JPEG Compression and Gaussian Smoothing have been used in various image processing applications and can be used as points of reference to our RSCD method.
\vspace{-0.4cm}
\subsubsection{{JPEG Compression.}}
JPEG compression is a widely-used technique that compresses the input data by sub-sampling the color information, quantizing the discrete cosines transform coefficients, and applying Huffman-Coding \cite{das2017keeping}.
We used the JPEG compression technique in \cite{clark2015pillow} to generate 95 different quality level images, in which scales 1 and 95 are the lowest and the highest quality levels, respectively.
We then selected a subset of 10 qualities for each image to match the CtF samples used in RSCD. More specifically, we selected $q \in \{1,2,4,6,8,12,16,32,64,95\}$, where $q$ is the quality scale. Figure \ref{fig:model_sample}(b) shows the selected quality levels of one image using the JPEG compression method.
\vspace{-0.4cm}
\subsubsection{Gaussian Smoothing.}
Gaussian smoothing is the method of smoothing an image using the Gaussian function to weigh the neighboring pixels \cite{xu2017feature}. Gaussian smoothing is widely used in image processing applications to reduce noise and other high-frequency details. We applied 10 different values for the standard deviation of the Gaussian kernel, $\sigma$, starting from 10 to 1, to  match the CtF samples used in RSCD and create 10 decomposed images. Figure \ref{fig:model_sample}(b) shows the selected quality levels of one image using the Gaussian smoothing method.

\subsection{Dataset}
In this study, we focus on image classification since it has become the leading task with a broad range of applications in machine learning and computer vision. Among popular datasets available for image classification, ImageNet has been widely used in evaluating cutting-edge models. Also, some prior studies have presented algorithms that could surpass human-level performance on this dataset \cite{russakovsky2015imagenet}. 
We used two datasets sub-sampled from ImageNet for our experiments. This approach allowed us to investigate the models' behavior with high resolution and diverse data compared to standard and smaller datasets such as CIFAR10 \cite{krizhevsky2009learning}. For the first set of experiments, including the off-the-shelf models' comparison and human subjects' results, we handpicked 10 classes from ImageNet that were visually distinctive. We refer to this subset as ImageNet10. 
We also randomly chose 20 classes of ImageNet, and added them to the 10 previously chosen classes, leading to 30 unique classes. We refer to this subset as ImageNet30. To study the scalability of our proposed algorithms and challenge existing models, the majority of our experiments were carried out on ImageNet30.
More specifically, we used a subset of the validation set of ILSVRC-2012\footnote{ILSVRC is the ImageNet Large Scale Visual Recognition Challenge \cite{russakovsky2015imagenet}.} which contains 50 images per class resized to $128\times128$. We generate 10 different versions of each image for our analyses leading to 15,000 images in total.

\subsection{Comparison of Deep Learning versus Human Perception} \label{subsec:motivation}
\subsubsection{Overconfident Deep Models.} We selected 4 off-the-shelf deep learning models with outstanding results on ImageNet, namely ResNet50 \cite{he2016deep}, ResNet152, InceptionV3 \cite{szegedy2016rethinking}, and Xception \cite{chollet2017xception}, and examined their performance on decomposed images of ImageNet10 over time. Based on the available neuroscience studies, we initially expected the deep models to perform with very low certainty and chance-level accuracy on the first time steps and quality levels and achieve higher confidence and accuracy over time as the image quality improves.
However, we observed that the deep models performed completely unexpectedly in the following cases:
\begin{itemize}
    \item[-] At early timesteps, low-quality and unintelligible images achieve confident but incorrect prediction as shown in Figures \ref{fig:machine_human}(a) and \ref{fig:machine_human}(d). Also, no steady increase is seen in the accuracy or confidence on higher quality images.
    \item[-] The classification accuracy, even on $Q_{10}$ images, is significantly less than the original images. Figure \ref{fig:machine_human}(a) shows deep models’ dependability to high-frequency information rather than the actual concepts.
    \item[-] Very spiky confidence and accuracy, even on high-quality images, due to models’ sensitivity to small and unnoticeable changes. See the magnified area in Figure \ref{fig:machine_human}(d), which shows ResNet50 spiky and overconfident behavior on $Q_{10}$. These spiky patterns are also noticeable even when models’ performance is averaged over all images in one class. See Figures \ref{fig:machine_human}(a), \ref{fig:machine_human}(b), \ref{fig:machine_human}(d), and \ref{fig:machine_human}(e).
    \item[-] Occasionally, the models' performance significantly drops at a higher quality. See qualities 6 and 7 in Figure  \ref{fig:machine_human}(b).
\end{itemize}
\vspace{-0.4cm}
\subsubsection{Different Visual Trajectory in Humans.} We conducted a similar task on human participants to verify our original hypothesis---confidence and accuracy increase over time---and compared humans' performance with that of deep learning models. In doing so, we asked seven volunteers to look at the images, type the main object they recognize in each image, and share their confidence level on their recognition. For each reconstruction method, one image was randomly selected from each of the ten previously mentioned classes. All different qualities of each image were shown to the participants, in order, starting from the lowest quality. We post-processed the participants’ answers to the basic level category \cite{callanan1985parents} employing the same structure used in defining class labels and hierarchies in creating ImageNet using WordNet lexical database \cite{fellbaum2010wordnet}. We then used the same categories for identifying the performance of the models.

\begin{figure}[!t]
 \centering
    \subfigure[ResNet50, Acc.]{\includegraphics[width=4cm]{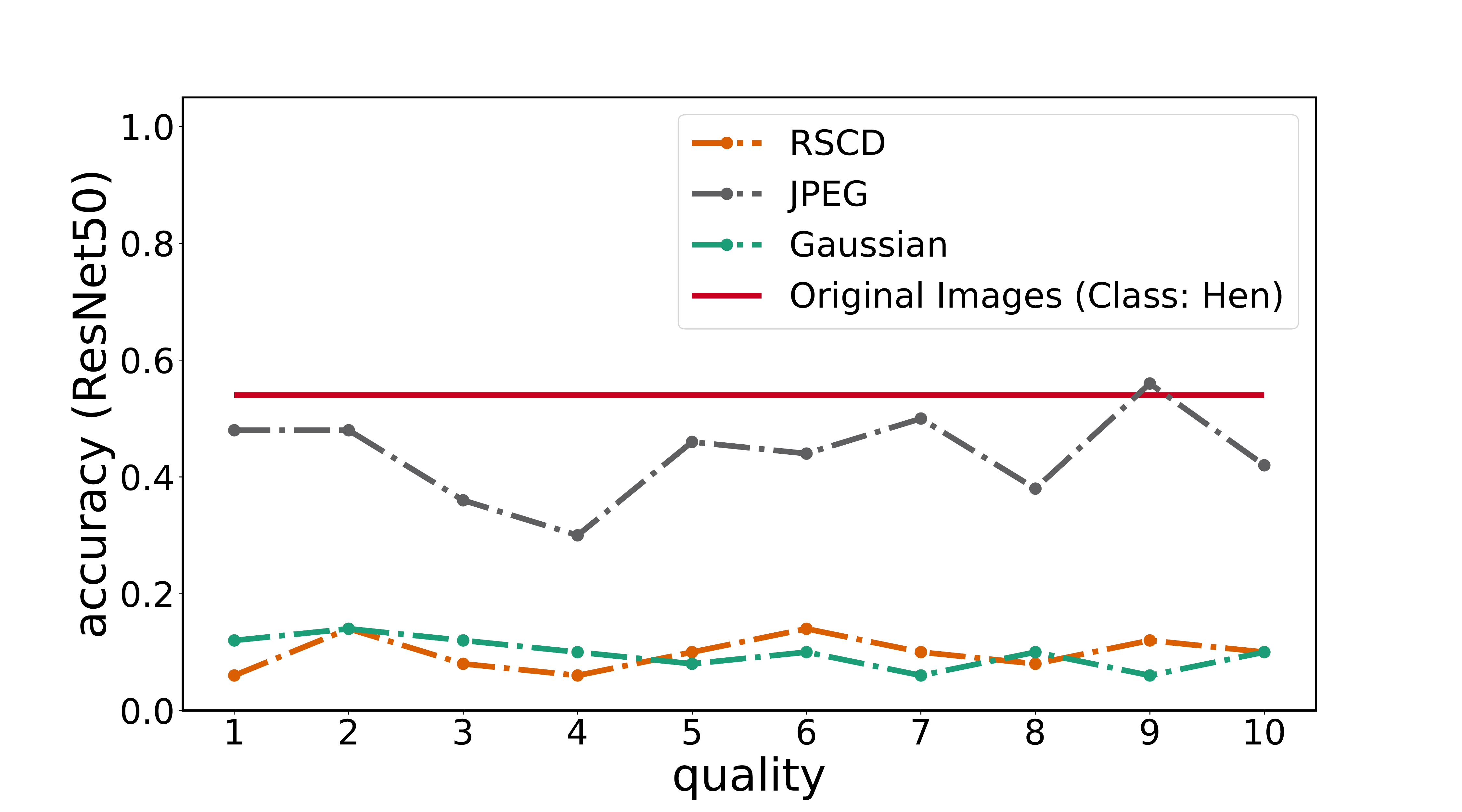}}
    \subfigure[DL, Acc.]{\includegraphics[width=4cm]{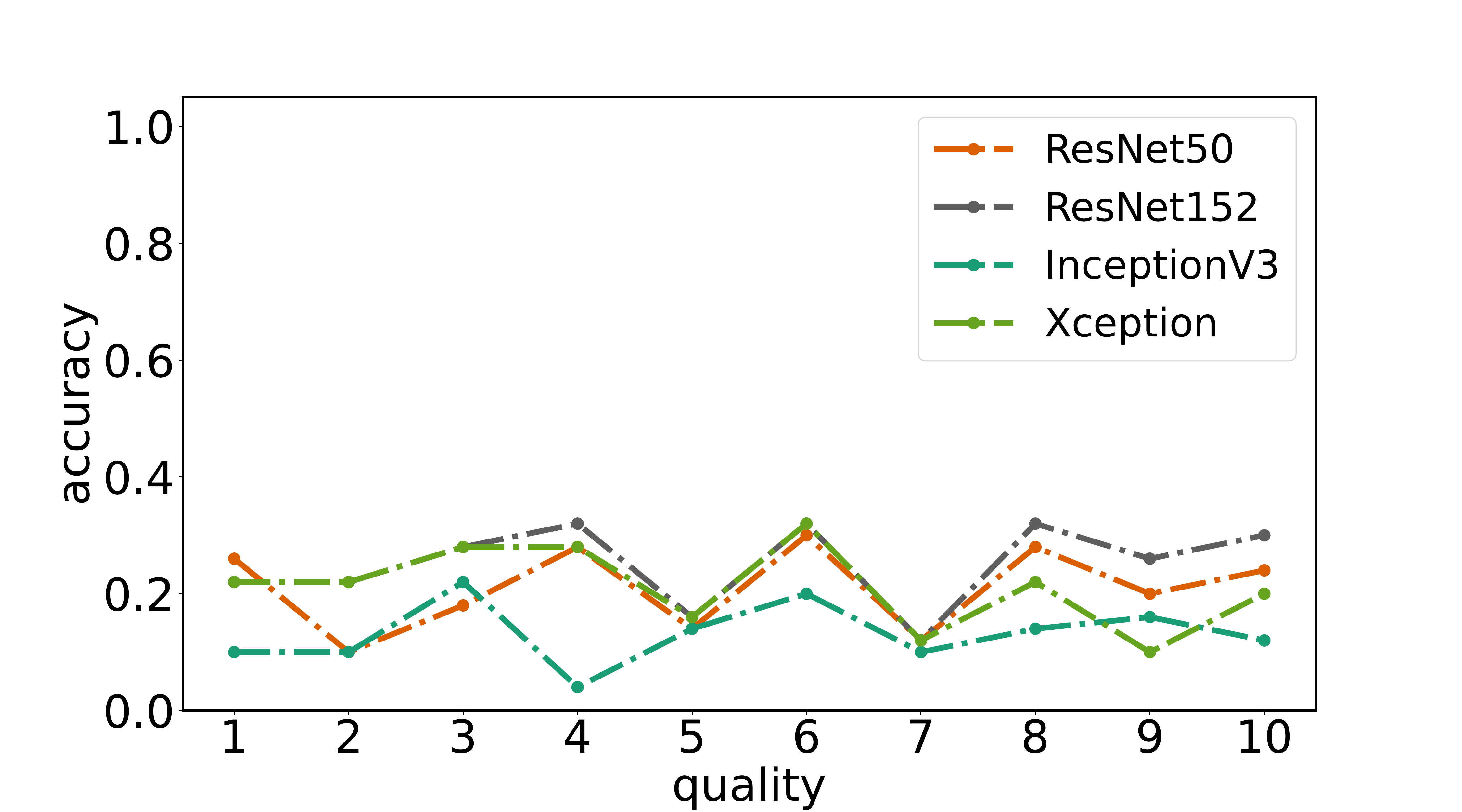}}
    \subfigure[Human vs. DL, Acc]{\includegraphics[width=4cm]{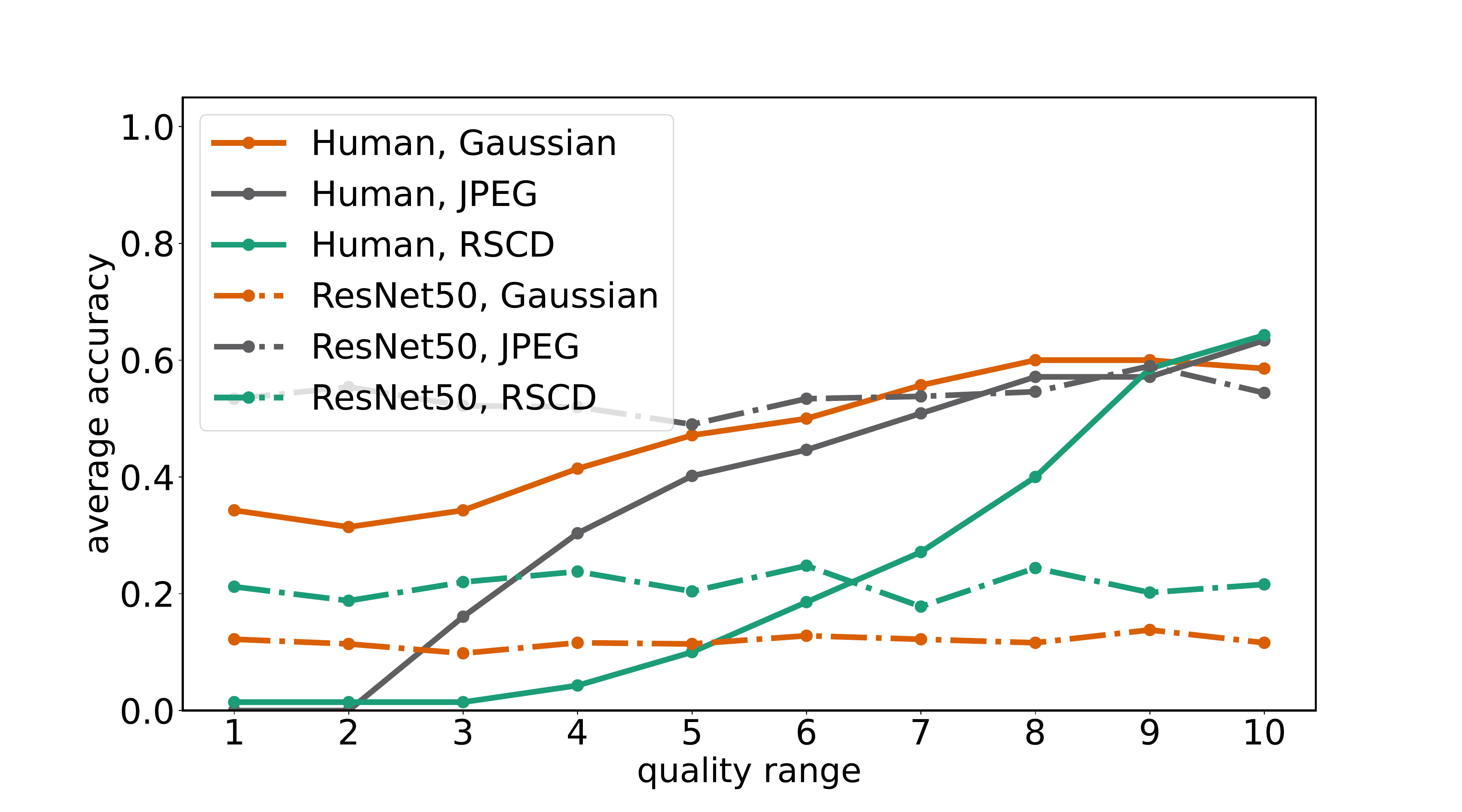}}
    \subfigure[ResNet50, Conf.]{\includegraphics[width=4cm]{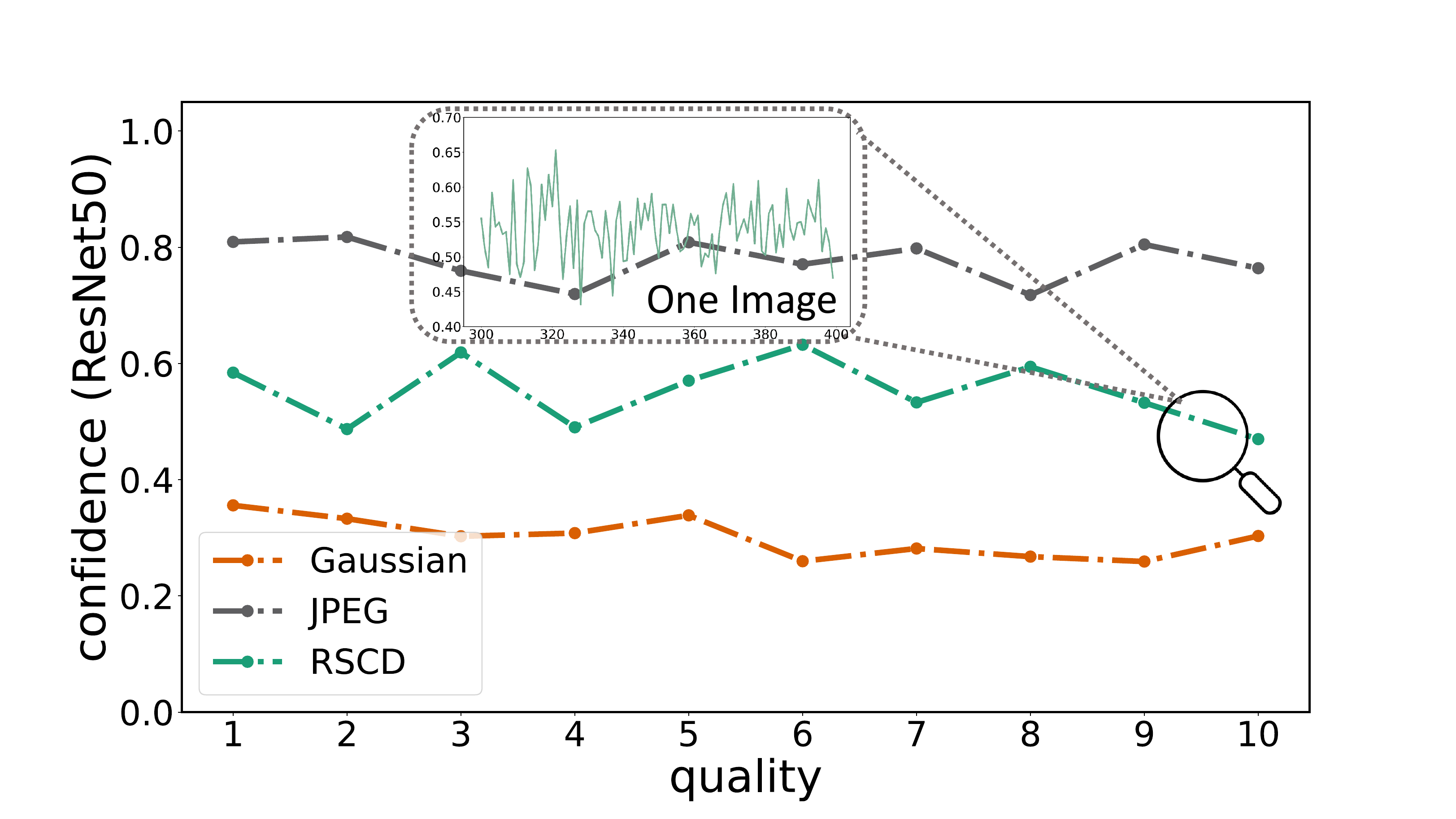}}
    \subfigure[DL, Conf.]{\includegraphics[width=4cm]{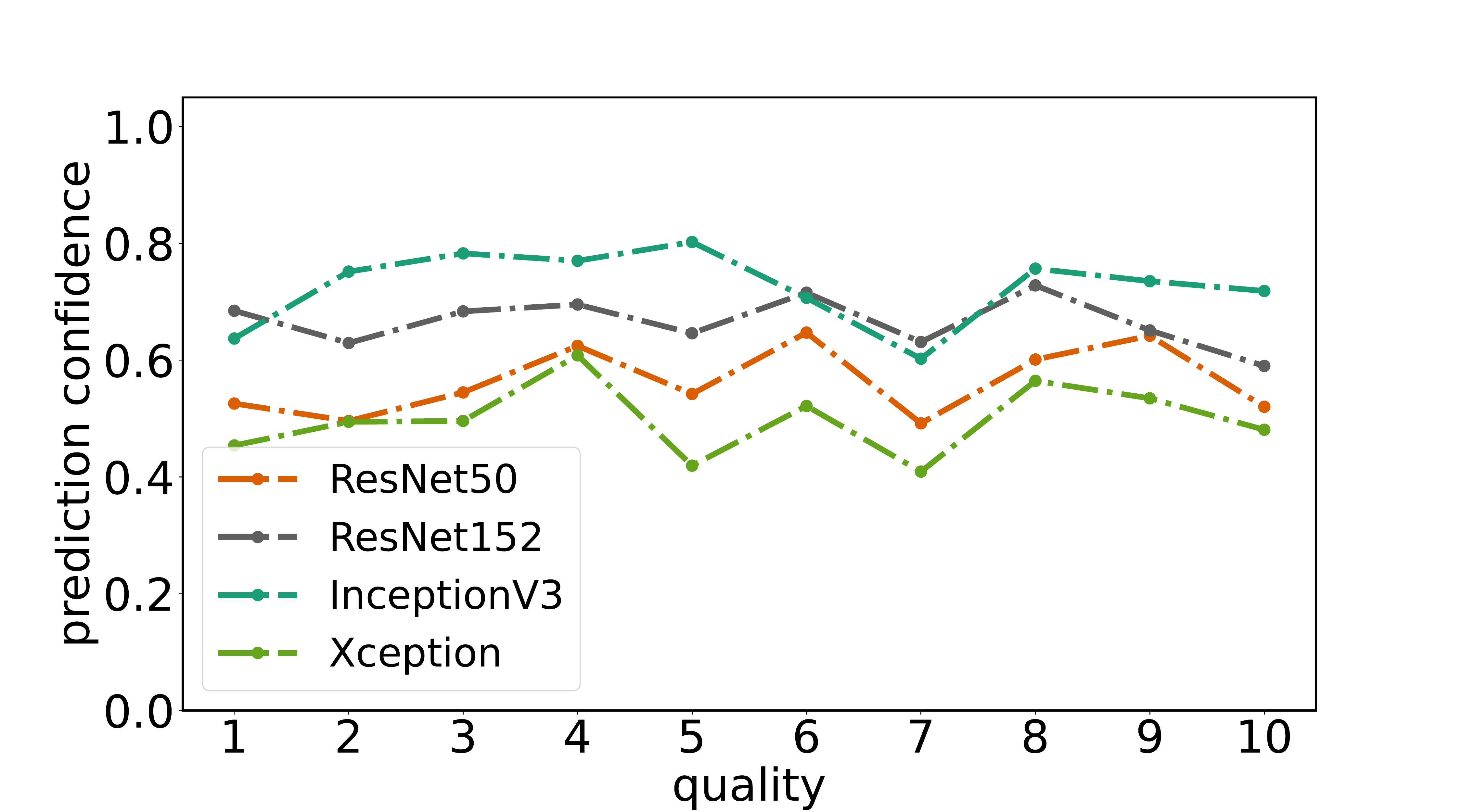}}
    \subfigure[Human vs. DL, Conf.]{\includegraphics[width=4cm]{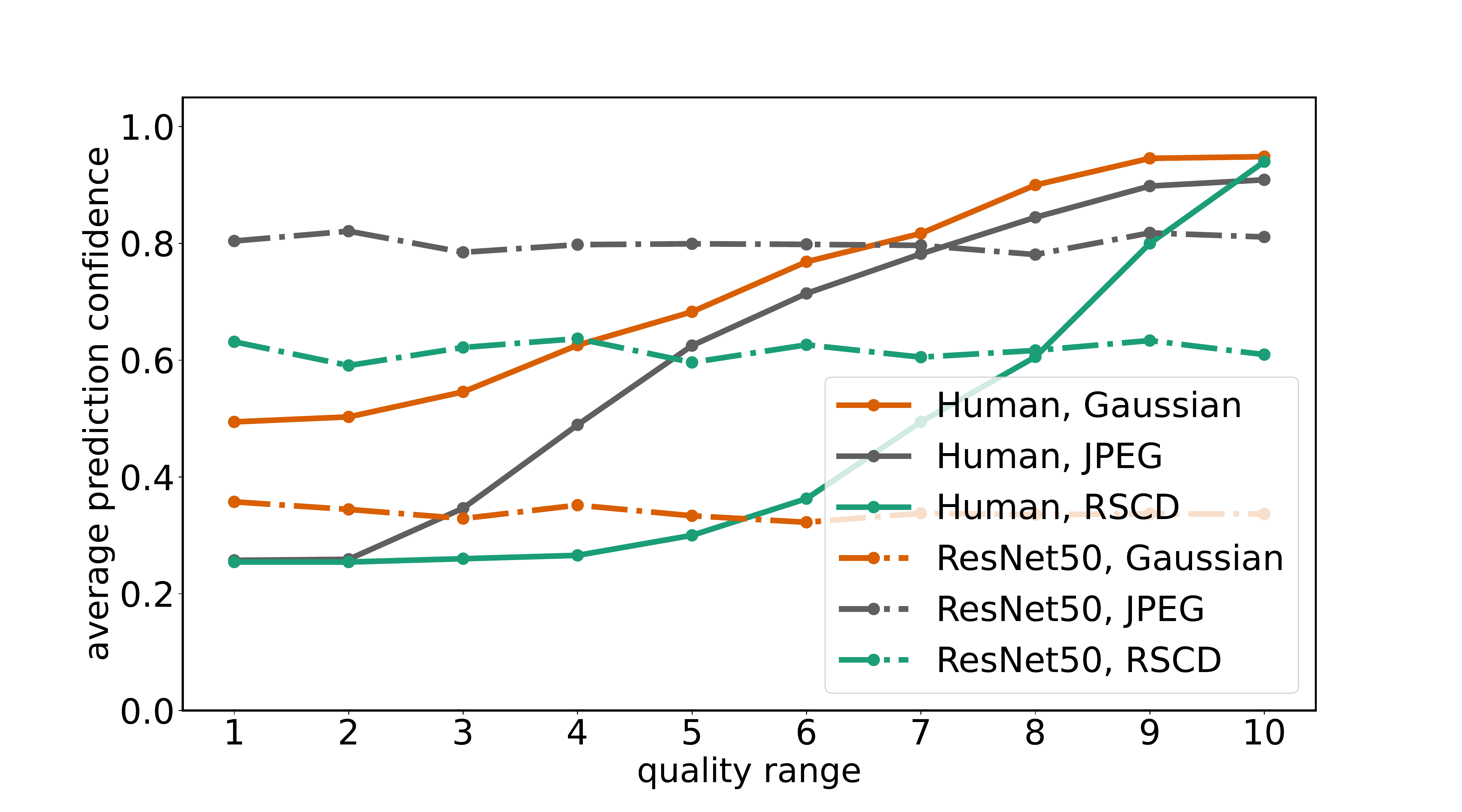}}

  \caption{Deep learning models are fragile and overconfident as they get better quality images. Humans seem to have a different trajectory in perception. (a),(d): ResNet50 performance on one class, namely, hen. The magnified area in (d) shows ResNet50's spiky confidence even for $t\in[300,400]$. (b),(e): Models' performance using RSCD on one class, namely, hen. (c),(d): Performance averaged over all classes in ImageNet10.}
  \vspace{-0.4cm}
  \label{fig:machine_human}
  \end{figure}

Our results show humans’ confidence and accuracy increase as the image quality increases. Unlike the deep learning models, there is almost no sudden change in human accuracy or confidence; instead, both accuracy and confidence increase gradually over time and quality. Figures \ref{fig:machine_human}(c) and \ref{fig:machine_human}(f) compare participants’ performance averaged over ImageNet10 classes with that of ResNet50 for all three decomposition methods.

These results motivated us to design a perception model that can close the gap between human results and deep models by taking advantage of CtF information. We also expect such a perception model to be more robust to minor perturbation and high-frequency data.
\vspace{-0.4cm}
\subsection{Our Solution for Perception Over Time}
Sequential models such as 1-D convolutional neural networks 
(1-D CNNs) and recurrent neural networks (RNNs) have shown interesting results in computer vision tasks that involve sequences such as video recognition. Some studies have incorporated such models into standard (2-D) CNNs and significantly improved the performance of recognition pipelines for sequential tasks \cite{donahue2015long,shi2015convolutional}.

Motivated by these studies and the different visual trajectories we saw in humans and machines in \ref{subsec:motivation}, we designed two sequential models, namely CtF-CNN and CtF-LSTM, that can take in information from static images over time and perform a more robust and accurate perception of them. In doing so, we use the introduced decomposition methods in \ref{subsec:decomposition_methods} to first generate a time component for static images and then input them in a CtF manner to the sequential models.
\vspace{-0.4cm}
\subsubsection{Models.} 
\begin{figure}[h]
 \centering
    \subfigure[Baseline Models]{\includegraphics[width=5cm]{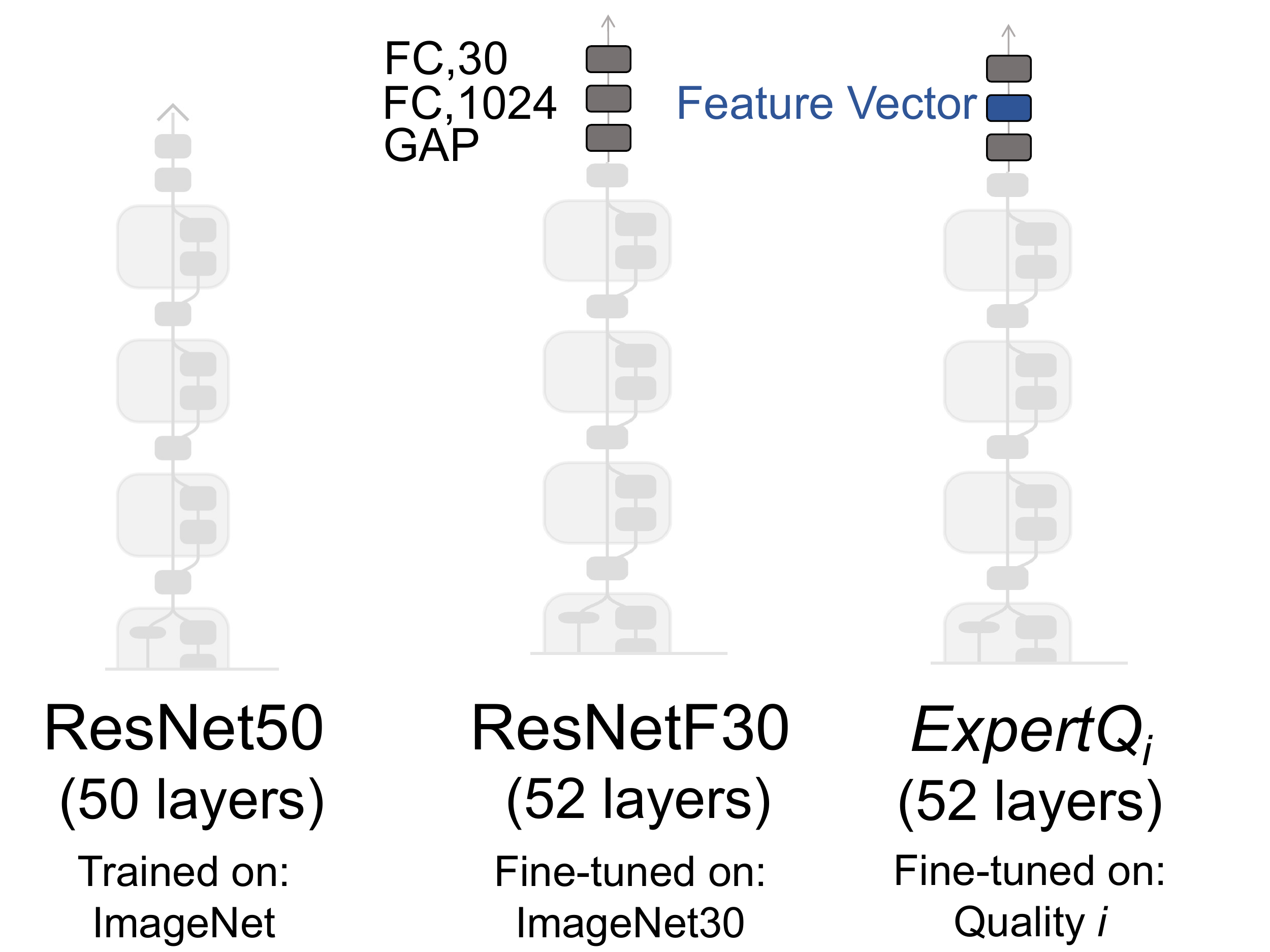}}
    \subfigure[CtF Model]{\includegraphics[width=7cm]{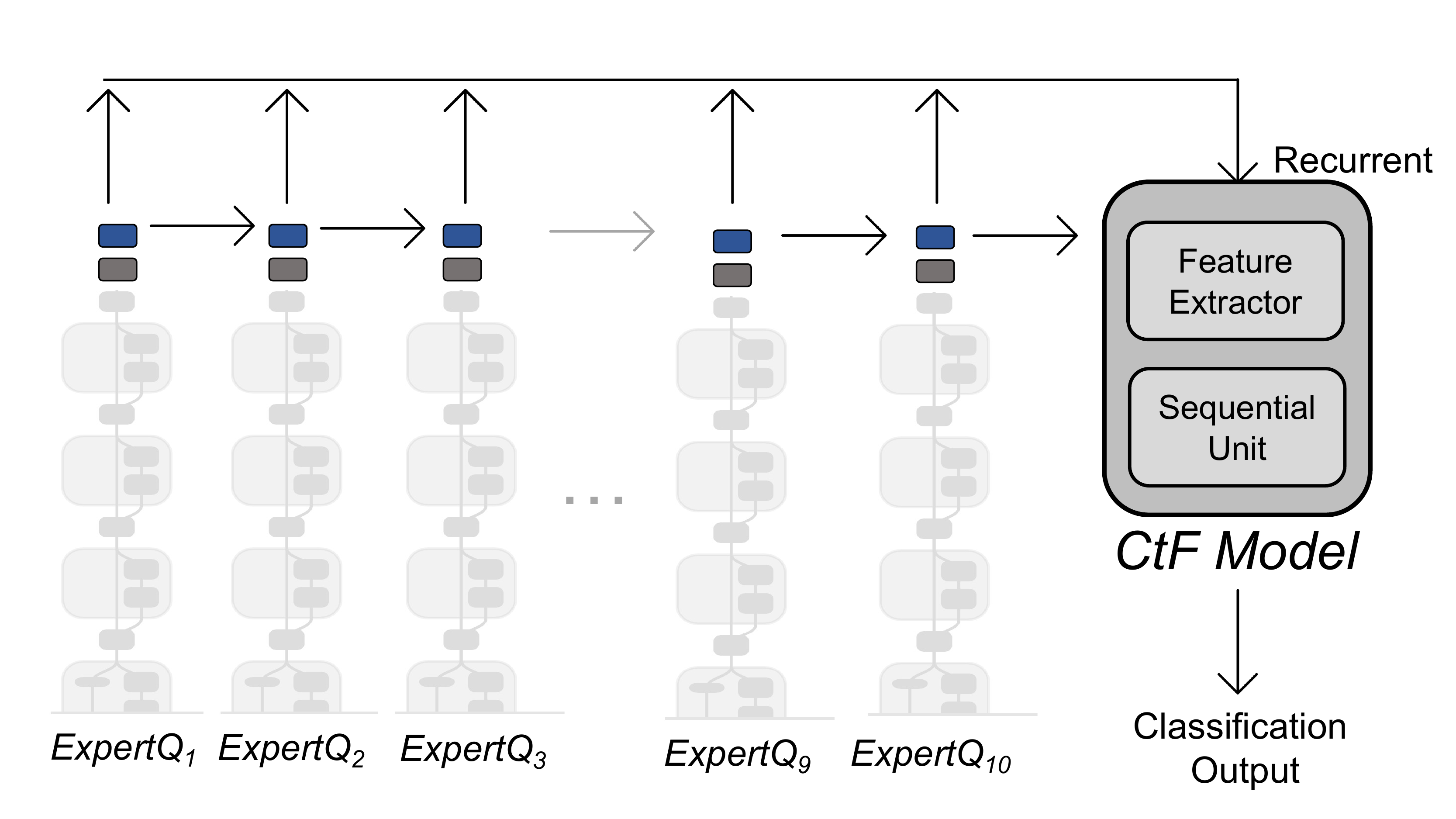}}
  \caption{Models architecture. ResNet50 illustration is borrowed from \cite{schubert_2020}.}
  
  \label{fig:models_architecture}
  \end{figure}

First, we  transferred the weights from ImageNet ResNet50 to a model and fine-tuned it on ImageNet30. We refer to this model as ResNetF30. More specifically, we removed the ResNet50 classification layer, added a global average pooling layer (GAP), and a fully connected layer (FC) with 1024 nodes, followed by a classification layer containing $N_{class}$ nodes, 30 in our case. See Figure \ref{fig:models_architecture}(a).

Secondly, for each decomposition quality level, we create a model based on ResNetF30 and fine-tune it on that quality level, leading to 10 Expert Models in total. We call these models $ExpertQ_{i}$, where $i$ is an integer representing the quality level of input decompositions and $i\in[1,10]$, and use them as baselines.

Furthermore, we take the final feature vector of each $ExpertQ_{i}$ -- the output of the layer prior to the classification layer-- as an input to the sequential models, CtF-CNN and CtF-LSTM, to process static images in a CtF manner and over time. See Figures \ref{fig:models_architecture}(a) and \ref{fig:models_architecture}(b) for the models architectures.

Our proposed sequential framework architecture has two main parts. The first part extracts features from the feature vectors received from Expert Models, and the second part performs the temporal perception by processing the sequences over time. This architecture allowed our models to extract meaningful features from the individual Experts and make the final classification decision based on all the information over time. In Section \ref{sec:experiments}, we show how such architectures make the classification models more accurate and robust.

\noindent \textbf{Model Architecture.}
As mentioned before, CtF-CNN and CtF-LSTM have a similar feature extraction part. This part takes the outputs of Expert Models, all or a selection of them, as a sequence and feeds them into two 1-D CNN layers, with 128 and 64 filters, respectively. The kernel size of the first 1-D CNN layer is equal to the size of the feature vector of each Expert Model, in our case is 1024, and the kernel size of the second 1-D CNN is half the size, in our case, 512.

For CtF-CNN, the feature extraction part is followed by three layers. A 1-D CNN with 32 filters and a kernel size of 256, followed by a global average pooling and a fully connected layer of size $N_{classes}$, where $N_{classes}$ is the number of classes in the dataset, i.e., 30 in ImageNet30.

For CtF-LSTM, the feature extraction part is followed by two layers. An LSTM layer with $N_{qualities} \times N_{classes}$ hidden cells---where $N_{qualities}$ is the number of selected Expert Models---and a fully connected layer of size $N_{classes}$. 

\noindent\textbf{Training.} During ResNetF30 fine-tuning, we used a Root Mean Squared Propagation optimizer with $\rho$ = 0.9, $\mu$ = 0.9, $\epsilon$ = 1e-07, and an initial learning rate of 0.001. We used L1 regularization with a regularization factor of 0.01, and trained the model for 35 epochs with a batch size of 10. For $ExpertQ_i$, we fine-tuned ResNetF30 on decomposed images of quality $i$ from the select decomposition method. See comparisons in Section \ref{sec:experiments}. We also used similar parameters used in training ResNetF30. In training CtF-CNN and CtF-LSTM, we applied batch normalization after each 1-D convolutional layer. Also, we employed an Adam optimizer with $\beta_1$ = 0.9, $\beta_2$ = 0.999 and an initial learning rate of 0.001. We trained both models for 35 epochs with a batch size of 10. Also, we used 75\% of the data for training and the rest for test in training all models.
\vspace{-0.4cm}
\subsubsection{Adversarial Robustness.}

Existing defense mechanisms for adversarial attacks in deep neural networks try to defend against the adversaries by augmenting the training data with adversarial examples \cite{madry2017towards}, or adding some level of stochasticity to the hidden layers \cite{das2017keeping}. Some studies also apply some preprocessing techniques to defend against such attacks \cite{xu2017feature,kim2019neuromorphic,guo2017countering}. It is worth mentioning that the main problem in deep neural networks is their tendency to learn surface-level predictability of the data rather than extracting concepts and meaningful information. As a result, while such techniques improve the models' performance to some extent, they do not help the models learn concepts. Thus, these models are still prone to high-frequency perturbation and attacks, which may not have been discovered yet. 

However, in our sequential models, we incorporate CtF visual processing observed in human perception with the hope of overcoming a broader set of high-frequency dependencies. We evaluate the robustness of the models by a gradient-based method, Projected Gradient Descent (PGD) \cite{nicolae2018adversarial}. PGD was motivated by previous studies as a universal first-order adversary that provides a guarantee against first-order attacks \cite{das2017keeping,kim2020modeling,madry2017towards}.

We attack ResNetF30 using PGD to the level that its prediction accuracy drops from 0.78 to 0.15. More specifically, we attack the model is an untargeted way, with $\epsilon$ = 5 and $\epsilon_{step}$ = 1 over 10 iterations. We collect the attacked images and run the decomposition methods to generate CtF attacked images. Our results demonstrate the effectiveness of our CtF visual perception models in the context of adversarial examples; however, we would like to emphasize that the implications of this model extend to robust perception in general. 

\begin{figure}[!t]
 \centering 
    \subfigure[Accuracy]{\includegraphics[width=4cm]{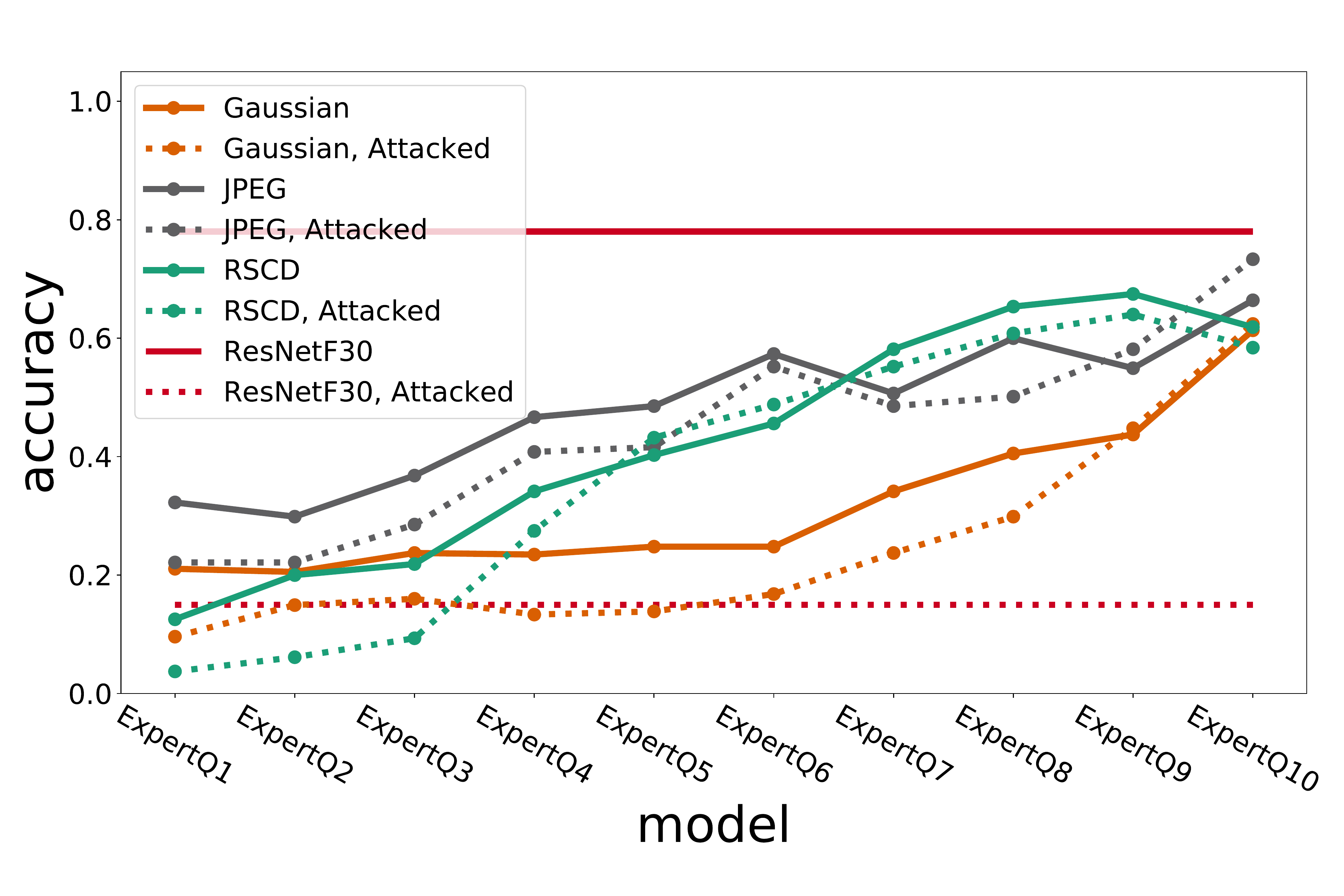}}
    \subfigure[Prediction Conf.]{\includegraphics[width=4cm]{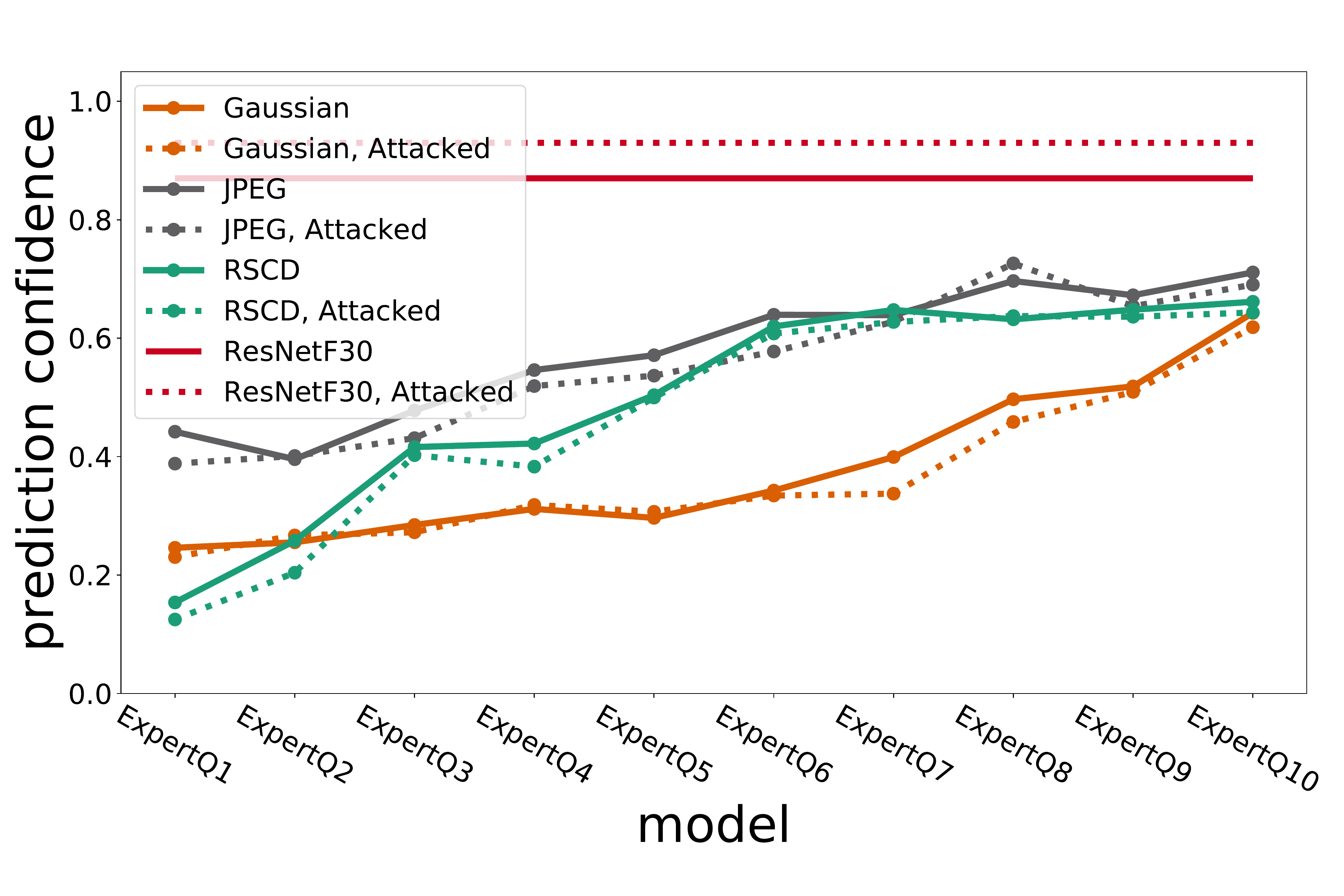}}
    \subfigure[True Class Conf.]{\includegraphics[width=4cm]{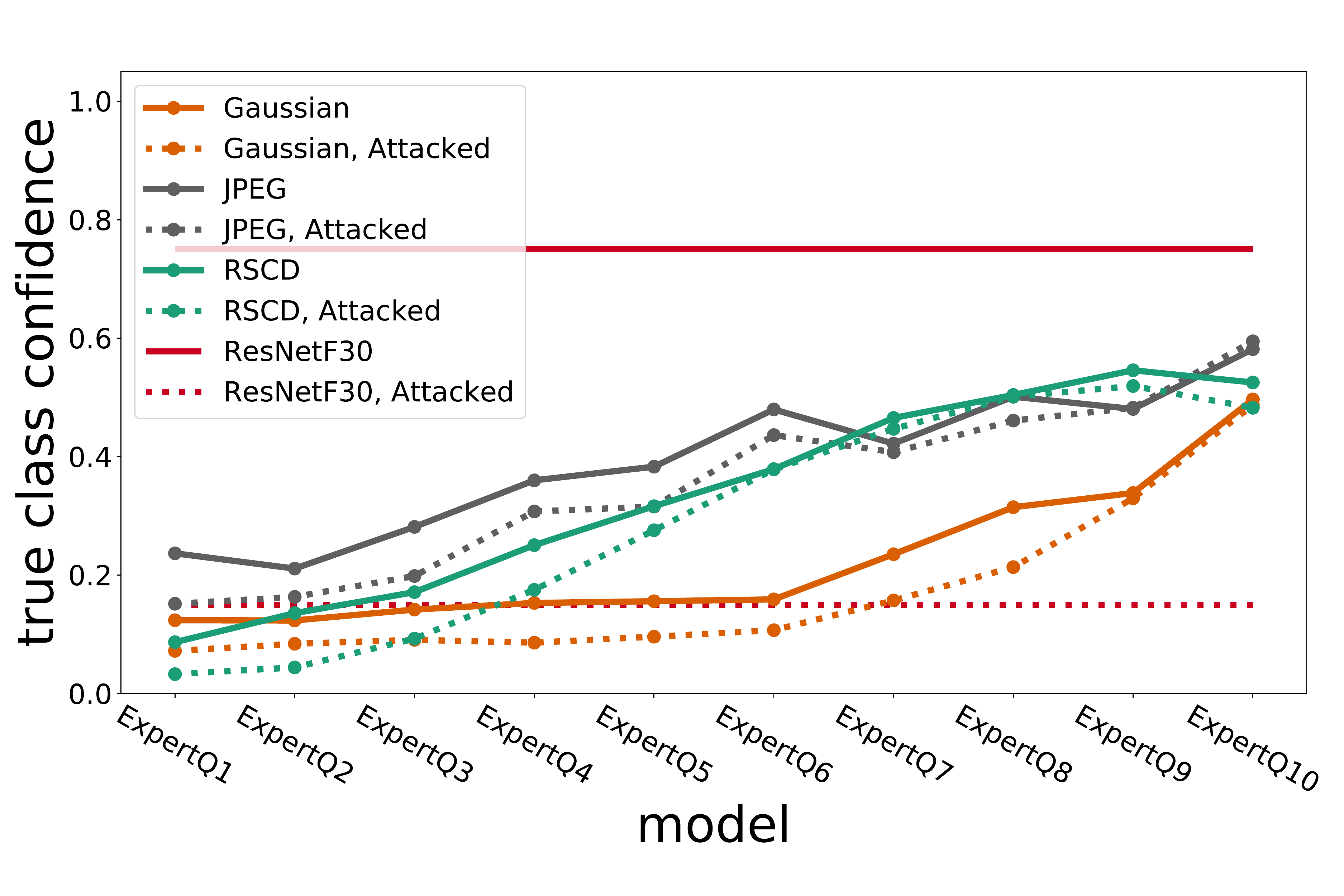}}
  \caption{Baseline models performance on decomposed images with different qualities.}
  \label{fig:baseline_experts}
  \end{figure}
\vspace{-0.5cm}
\section{Experiments and Results} \label{sec:experiments}
\subsection{Classification Accuracy}
To study the performance of the models, we compare their classification accuracy on the decomposed test images using all three decomposition methods. We refer to the performance of the models on unperturbed decomposed images as the standard performance, i.e., Acc., and refer to their performance on the adversarial perturbed images as ``attacked'', i.e., Att.

\vspace{-0.3cm}
\subsubsection{Baseline.} For the baseline analysis, we looked at the standard performance of ResNetF30 as well as Expert Models. See solid lines in Figure \ref{fig:baseline_experts}. For all decomposition methods, $ExpertQ_i$ generally performed better than $ExpertQ_j$, where $j < i$. The increasing pattern in the accuracy is in line with previous studies using JPEG compression and Gaussian smoothing \cite{das2017keeping,kim2020modeling}. Note that: 1.We did not see such an increasing pattern when we examined deep learning models trained on the original images, see Figures \ref{fig:machine_human}(b) and \ref{fig:machine_human}(e). 2. The decomposed images of quality $Q_{10}$ are not fully reconstructed and have a visually noticeable difference from the original images, especially in RSCD and Gaussian smoothing. Thus, the standard accuracy of ResNetF30 is higher than $ExpertQ_{10}$. However, $ExpertQ_{10}$ is relatively less overconfident of its detection.

Furthermore, Expert Models perform significantly better on the adversarial test images. See the dashed lines in Figure \ref{fig:baseline_experts}. Similar to the standard performance, we see an increasing pattern in the attacked performance of Expert Models. We also see that Expert Models are significantly less overconfident on the adversarial images compared to ResNetF30.
\vspace{-0.3cm}
\subsubsection{CtF Models.}
 We are interested in evaluating the key role of CtF processing in visual perception.
 In doing so, we compare the accuracy of the baseline models, ResNetF30 and $ExpertQ_{10}$ (EQ10), with that of our CtF models on unperturbed (Acc.) and adversarial attacked (Acc. Att.) images. See Table \ref{tab:accuracy_comparison} for the comparisons. Note that CtF-CNN and CtF-LSTM have access to all level decomposed images for their classification. These led to the following two impactful results. 

\begin{enumerate}
     \item \textbf{Our results show transformative accuracy improvements, almost perfect accuracy and over 20\% jump in accuracy over a fine-tuned ResNet50 model (ResNetF30) on the same dataset.}
     \item \textbf{Our CtF models on attacked RSCN (0.768) and JPEG decomposition (0.831) perform on par (at times better) than a standard ResNet model even when tested on non-attacked images (0.78). The use of our models results in no penalty to accuracy on unperturbed images.}
 \end{enumerate}
 
\begin{table}[t]
\caption{Accuracy comparisons between baselines and our CtF models.}
\vspace{-0.2cm}
\label{tab:accuracy_comparison}
\resizebox{\textwidth}{!}{%
\begin{tabular}{lllllllllll}
 &  & \multicolumn{3}{c}{EQ10} & \multicolumn{3}{c}{CtF-CNN} & \multicolumn{3}{c}{CtF-LSTM} \\ \hline
\multicolumn{1}{l|}{\textbf{Model}} & \multicolumn{1}{l|}{ResNetF30} & Gaussian & JPEG & \multicolumn{1}{l|}{RSCD} & Gaussian & JPEG & \multicolumn{1}{l|}{RSCD} & Gaussian & JPEG & RSCD \\ \hline
\multicolumn{1}{l|}{\textbf{Acc.}} & \multicolumn{1}{l|}{0.78} & 0.613 & 0.663 & \multicolumn{1}{l|}{0.618} & 0.922 & \textbf{0.989} & \multicolumn{1}{l|}{\textbf{0.989}} & 0.861 & 0.861 & 0.978 \\
\multicolumn{1}{l|}{\textbf{Acc. Att.}} & \multicolumn{1}{l|}{0.15} & 0.624 & 0.733 & \multicolumn{1}{l|}{0.583} & 0.600 & \textbf{0.831} & \multicolumn{1}{l|}{0.768} & 0.547 & 0.547 & 0.663 \\ \hline
\end{tabular}%
}
\end{table}

\subsection{Ablation Experiments}
\begin{figure}[!t]
 \centering
    \subfigure[Gaining Data]{\includegraphics[width=5.8cm]{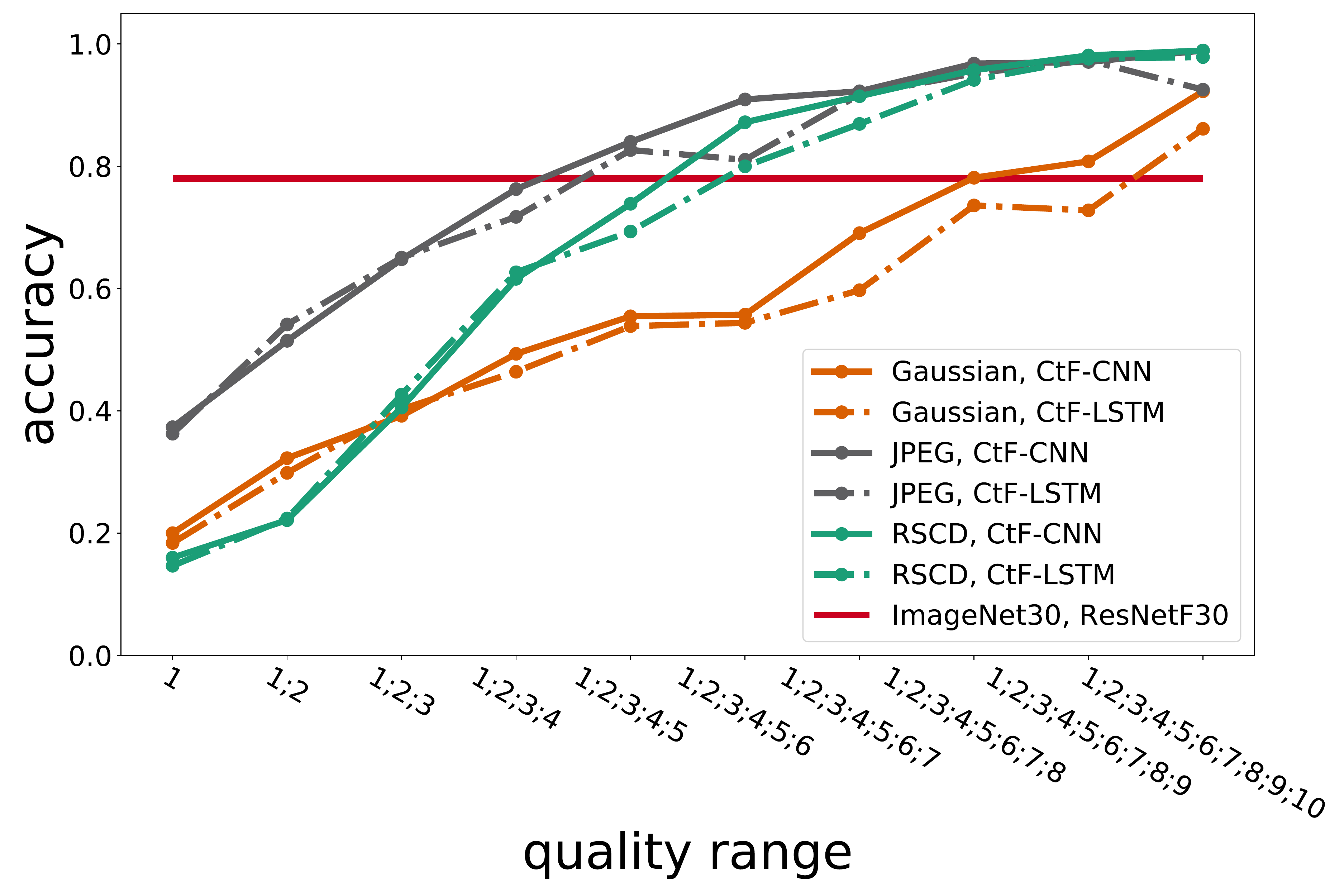}}
    \subfigure[Losing Data (Y-axis starts from 0.5)]{\includegraphics[width=6.2cm]{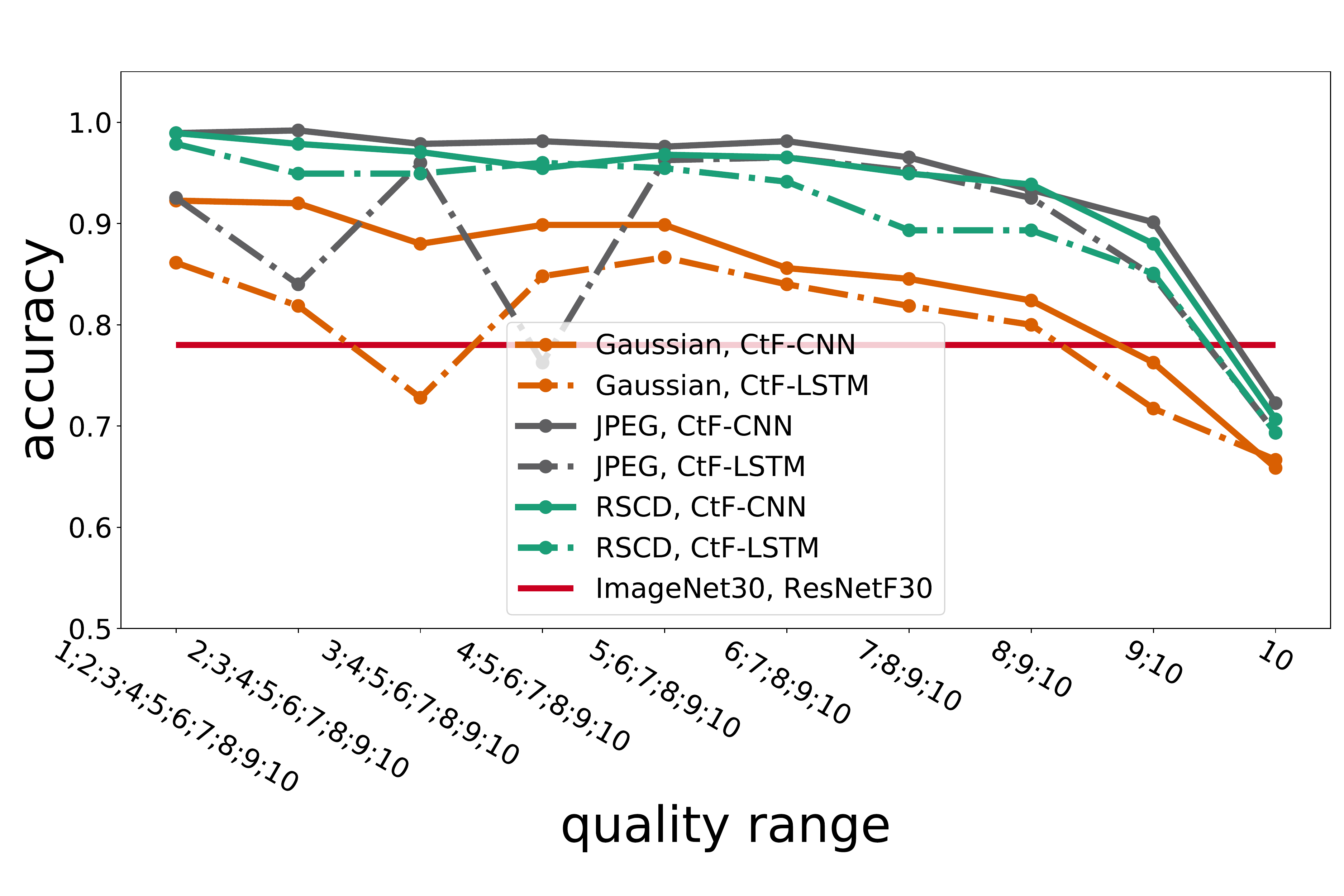}}
  \caption{ CtF models performance on ``gaining'' and ``losing'' data.}
  \label{fig:gaining_losing}
\end{figure}

For the ablation experiments, we studied the role of each quality level information and evaluated CtF-CNN and CtF-LSTM on two main categories: 1. ``Gaining data'': receiving better quality images, one quality level at a time, until all qualities are seen 2. ``Losing data'': skipping the lowest quality images, one quality level at a time, until only $Q_{10}$ is left. These categories allow us to discover the role of each quality level while interacting with other qualities in visual perception. Our results demonstrate that the standard performance of CtF-CNN and CtF-LSTM improves while ``gaining'' data, see Table \ref{tab:gaining_data}, which emphasizes the importance of fine-level input. Also, our proposed models outperform ResNet30 using any of the decomposition methods. See Figure \ref{fig:gaining_losing}(a). We demonstrate the importance of coarse-level data in visual perception, as shown in Figure \ref{fig:gaining_losing}(b). We see a general decreasing pattern in the accuracy of both models on all decomposition methods, even when losing $Q_1$ data. We also see a sudden accuracy drop at level 3 for Gaussian smoothing and at levels 2 and 4 for JPEG compression on the adversarial images. We believe this is caused by the poor ability of these models in approximating decomposition with no artifacts.
One other key finding here is that CtF models outperform ResNet30 even by using a limited quality level data, for example, with the quality range of [8,9,10]. 

\begin{table}[!t]
\centering
\caption{CtF models surpass ResNetF30 while ``gaining'' data. Cells with the minimum quality range that top ResNetF30 (0.78) are marked.}
\vspace{-0.3cm}
\label{tab:gaining_data}
\begin{tabular}{lcccccc}
 & \multicolumn{3}{c}{\textbf{CtF-CNN, Acc.}} & \multicolumn{3}{c}{\textbf{CtF-LSTM, Acc.}} \\ \cline{2-7} 
\textbf{Quality Range} & \multicolumn{1}{l}{Gaussian} & \multicolumn{1}{l}{JPEG} & \multicolumn{1}{l|}{RSCD} & \multicolumn{1}{l}{Gaussian} & \multicolumn{1}{l}{JPEG} & \multicolumn{1}{l}{RSCD} \\ \hline
\multicolumn{1}{l|}{{[}1{]}} & 0.200 & 0.373 & \multicolumn{1}{c|}{0.160} & 0.184 & 0.363 & 0.147 \\
\multicolumn{1}{l|}{{[}1; 2{]}} & 0.323 & 0.515 & \multicolumn{1}{c|}{0.221} & 0.299 & 0.541 & 0.224 \\
\multicolumn{1}{l|}{{[}1; 2; 3{]}} & 0.392 & 0.648 & \multicolumn{1}{c|}{0.405} & 0.403 & 0.651 & 0.427 \\
\multicolumn{1}{l|}{{[}1; 2; 3; 4{]}} & 0.493 & 0.763 & \multicolumn{1}{c|}{0.616} & 0.464 & 0.717 & 0.627 \\ \cline{3-3} \cline{6-6}
\multicolumn{1}{l|}{{[}1; 2; 3; 4; 5{]}} & 0.555 & 0.840 & \multicolumn{1}{c|}{0.739} & 0.539 & 0.827 & 0.693 \\ \cline{4-4} \cline{7-7} 
\multicolumn{1}{l|}{{[}1; 2; 3; 4; 5; 6{]}} & 0.557 & 0.909 & \multicolumn{1}{c|}{0.872} & 0.544 & 0.811 & 0.800 \\
\multicolumn{1}{l|}{{[}1; 2; 3; 4; 5; 6; 7{]}} & 0.691 & 0.923 & \multicolumn{1}{c|}{0.915} & 0.597 & 0.917 & 0.869 \\ \cline{2-2}
\multicolumn{1}{l|}{{[}1; 2; 3; 4; 5; 6; 7; 8{]}} & 0.781 & 0.968 & \multicolumn{1}{c|}{0.957} & 0.736 & 0.952 & 0.941 \\
\multicolumn{1}{l|}{{[}1; 2; 3; 4; 5; 6; 7; 8; 9{]}} & 0.808 & 0.971 & \multicolumn{1}{c|}{0.981} & 0.728 & 0.973 & 0.976 \\ \cline{5-5}
\multicolumn{1}{l|}{{[}1; 2; 3; 4; 5; 6; 7; 8; 9; 10{]}} & 0.923 & 0.989 & \multicolumn{1}{c|}{0.989} & 0.861 & 0.925 & 0.979 \\ \hline
\end{tabular}%
\end{table}

\subsection{Adversarial Robustness}
\begin{figure}[!t]
    \centering
    \subfigure[RSCD, Gaining]{\includegraphics[width=4cm]{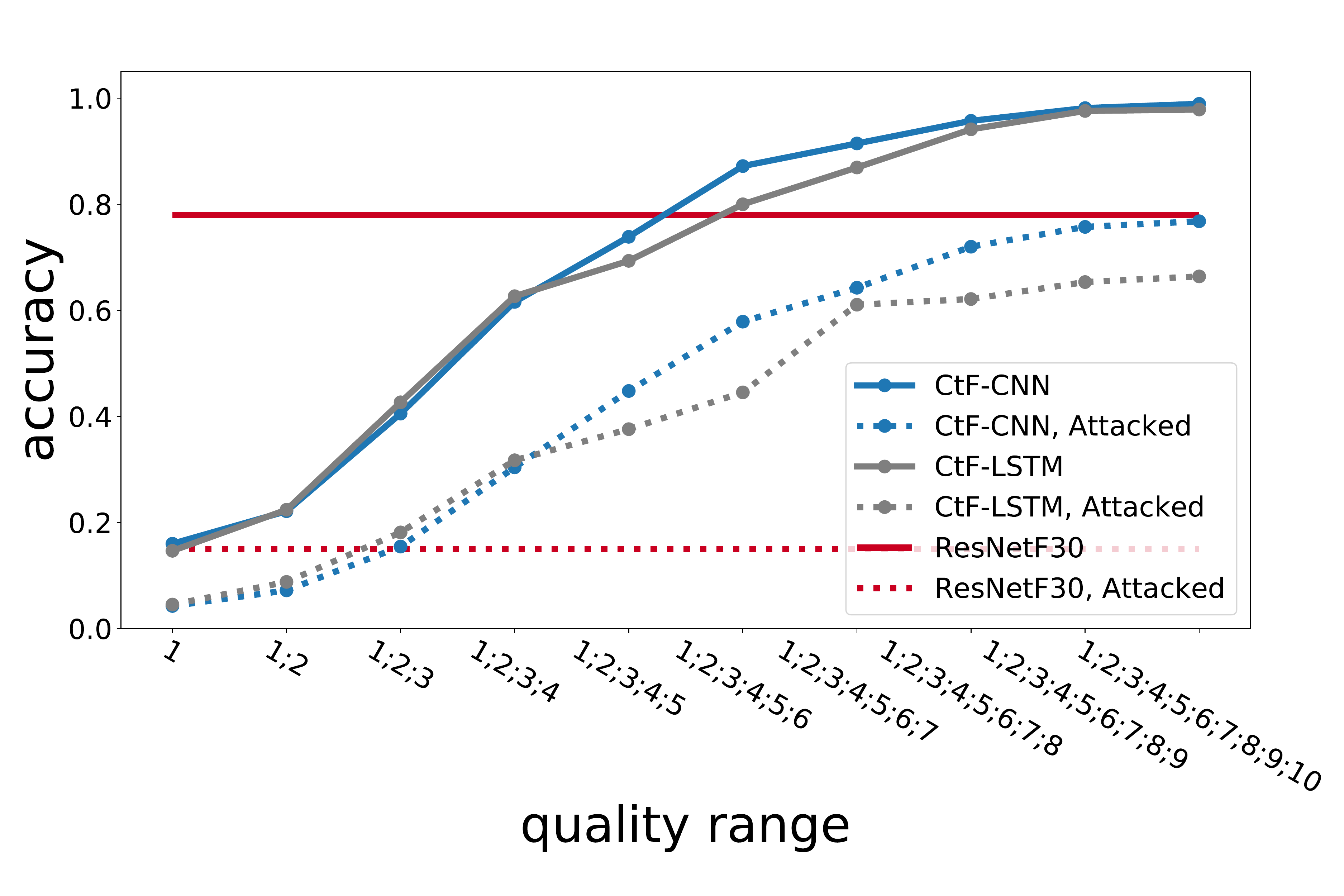}}
    \subfigure[JPEG, Gaining]{\includegraphics[width=4cm]{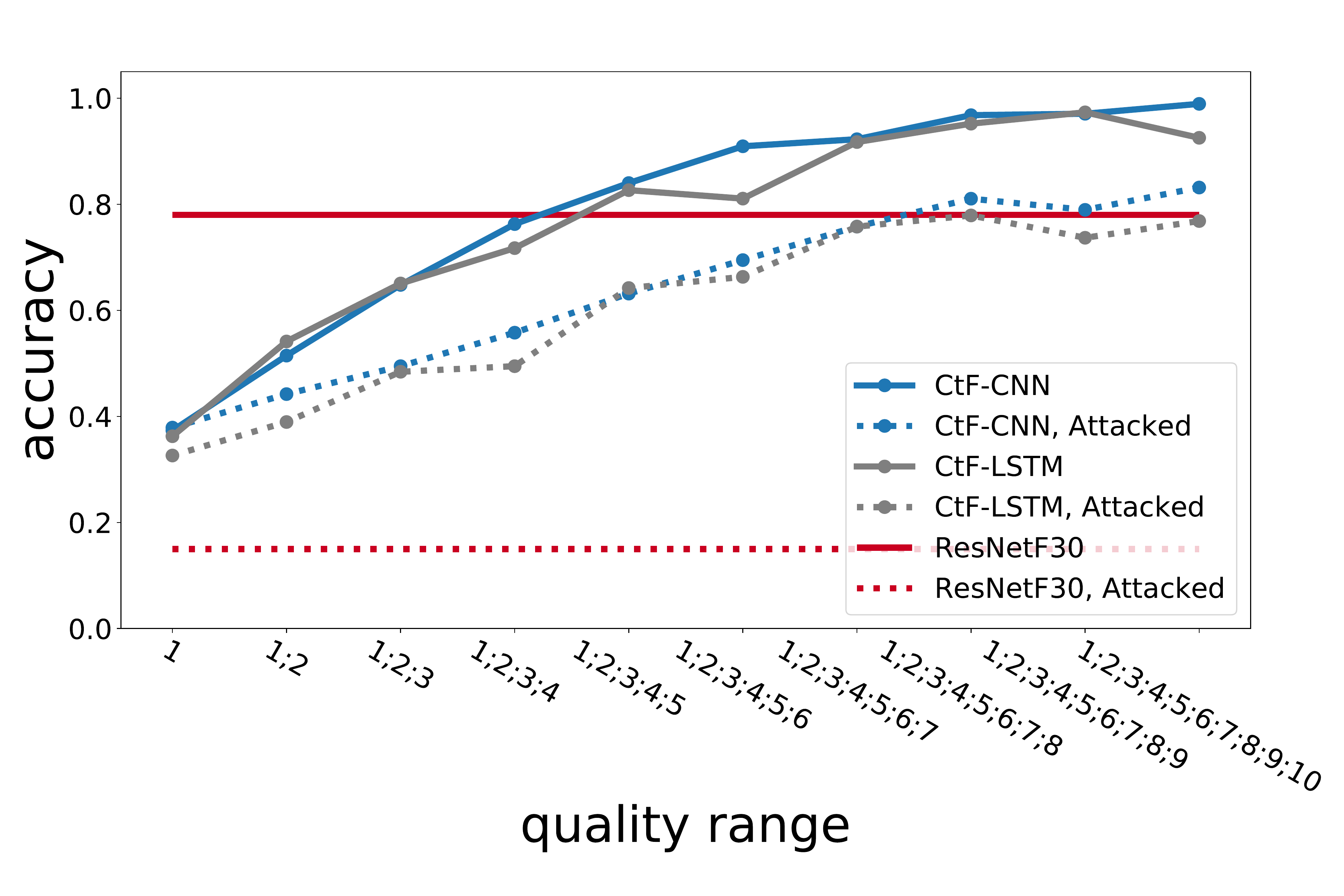}}
    \subfigure[Gaussian, Gaining]{\includegraphics[width=4cm]{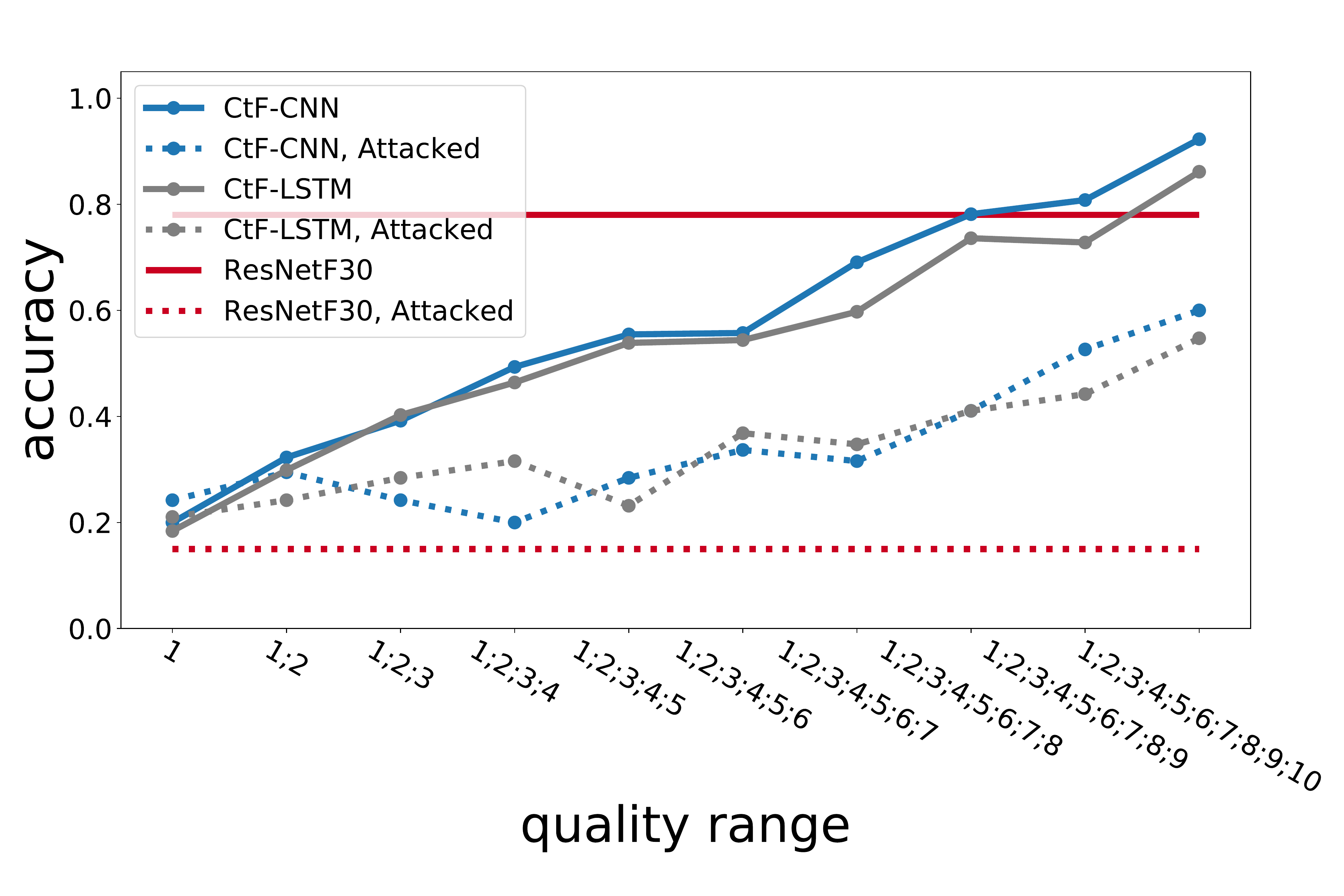}}    
    \subfigure[RSCD, Losing]{\includegraphics[width=4cm]{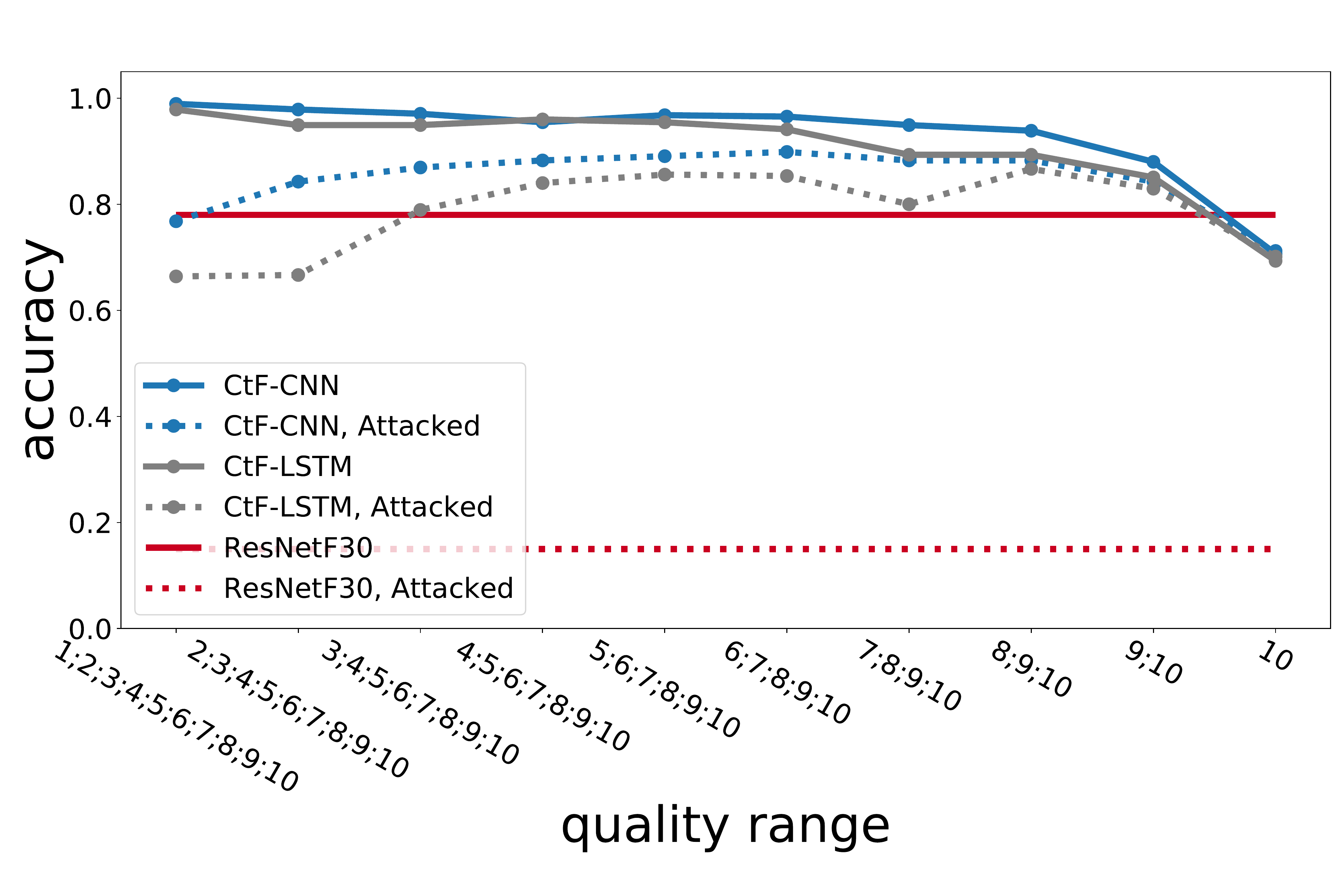}}
    \subfigure[JPEG, Losing]{\includegraphics[width=4cm]{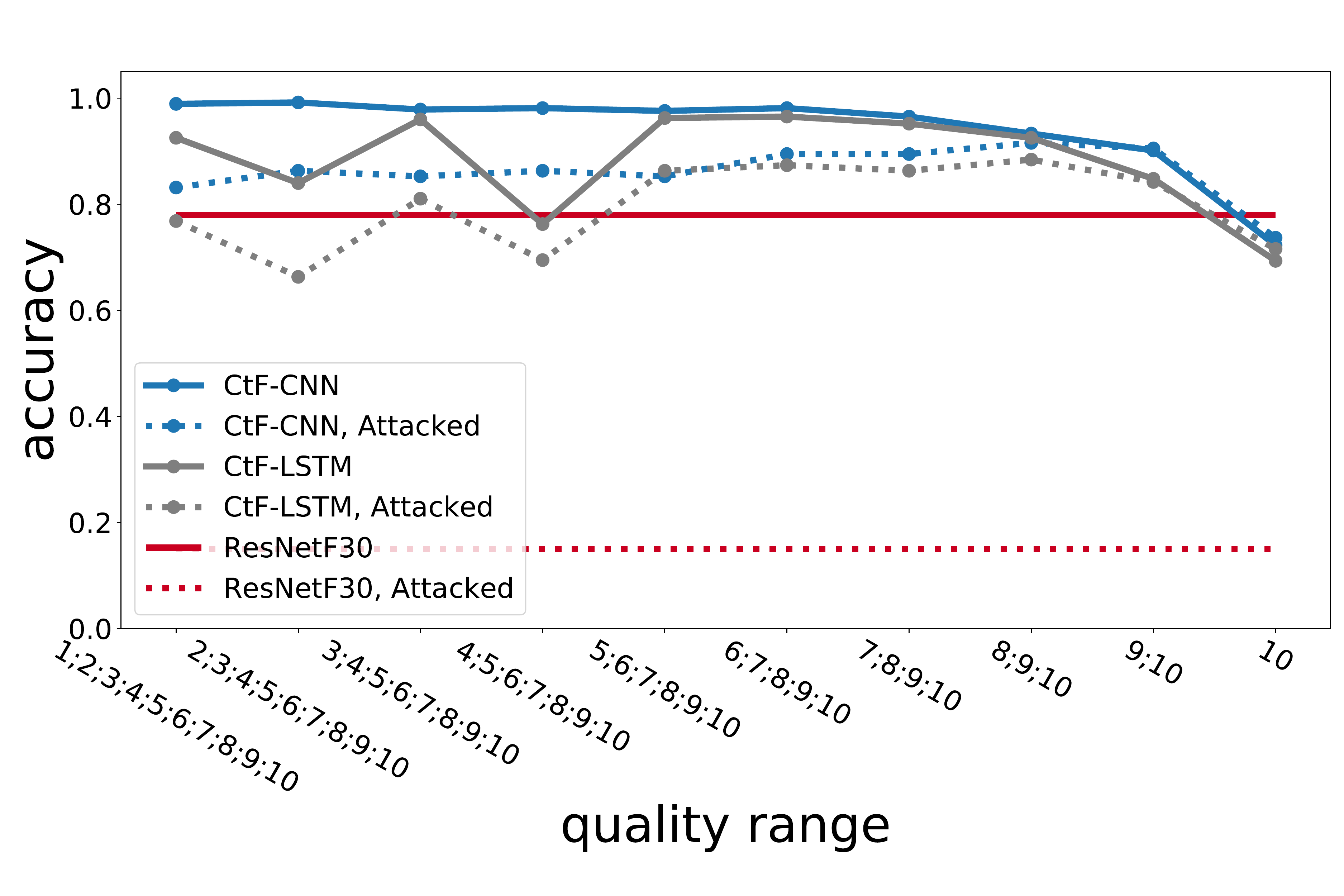}}
    \subfigure[Gaussian, Losing]{\includegraphics[width=4cm]{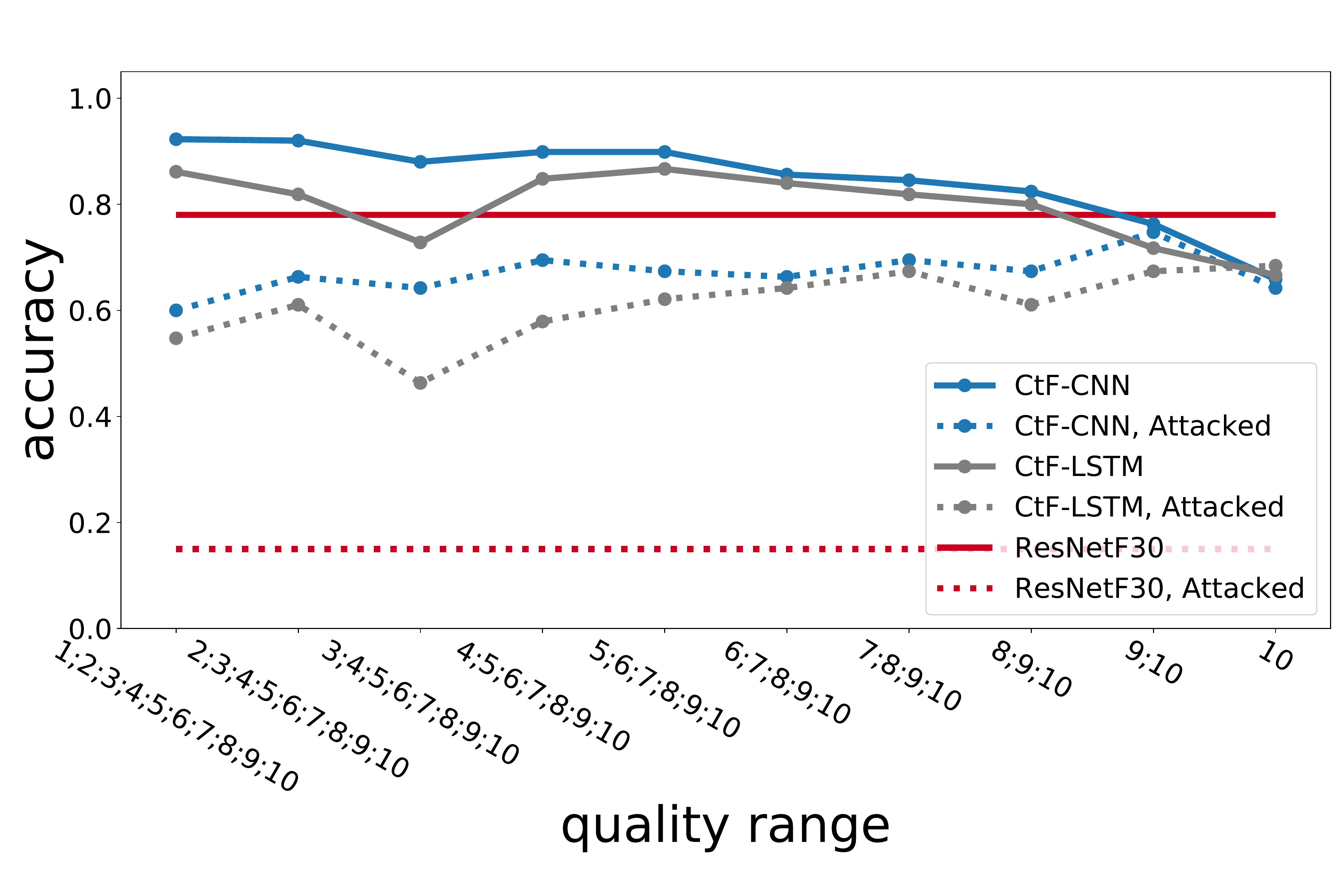}}
    \caption{Performance of the decomposition methods on ``gaining'' and ``losing'' data.}
    \label{fig:gaining_losing_attacked}
\end{figure}

In addition to the adversarial study in Table \ref{tab:accuracy_comparison}, we evaluated CtF-CNN and CtF-LSTM performance on the adversarial images while ``gaining'' and ``losing'' data. Our results show an increasing pattern in the accuracy of both models on all decomposition methods when gaining data, even on adversarial attacked images. See dashed lines in Figures \ref{fig:gaining_losing_attacked}(a), \ref{fig:gaining_losing_attacked}(b), and \ref{fig:gaining_losing_attacked}(c). We also see that our proposed decomposition, RSCD, is more robust to changes in the quality range. In contrast to JPEG compression and Gaussian smoothing, there is no sudden drop in the accuracy when losing one quality data using RSCD. 

\vspace{-0.4cm}
\subsection{Limitations}
Two main limitations of our proposed method are increased computation and memory costs.  The course-to-fine decomposition methods are required to be generated for each image before inference.  Then if the model is computed sequentially, each additional time step in inference will result in an additional forward pass through the model.  Furthermore, memory will be required to store each of the timesteps/model parameters and results.  

While it is true that the memory requirements are increased, we counter that there are minimal computational downsides to performing our approach on image recognition tasks.  In fact, with neuromorphic hardware \textit{even orders of magnitude computational and energy efficiency can be achieved over standard von Neumann implementations}.  First, the CtF images are not handled sequentially, but rather in a single pass through the model (similar to batching \cite{justus2018predicting} ), resulting in only a minimal computational percentage increase proportional to the input size. 

Furthermore, our RSCD model can be very efficiently computed in both time and energy costs on spiking neural network hardware.  In fact, Davies et al. \cite{davies2018loihi} have shown significant improvements in sparse coding both in energy consumption and speed. As an example, Loihi is a neuromorphic manycore processor with on-chip learning. It implements a spiking neural network in silicon and demonstrated LCA on their architecture with an improvement by \textit{48.7$\times$ energy consumption, 118$\times$ delay, and 5760$\times$ EDP (energy delay product)} over a CPU implemented FISTA (fast iterative shrinking-thresholding algorithm).

\vspace{-0.4cm}
\section{Conclusions}

Static image classification forms the basis of many computer vision problems, including but not limited to machine vision, medical imaging, autonomous vehicles, and more. Many of these areas are not immune to perturbations and require better trust and safety guarantees. 
Many studies have proposed defense mechanisms that were eventually found to be vulnerable, and our work may not be an exception to that. However, our goal is to get inspiration from human perception to build a more robust static image understanding. We find processing over time missing in many applications and see that as a key to misleading immunity.

As more information comes in over time, a model should become more and more confident of its conclusion.  However, current deep learning models do not have this property.  Rather small changes in the input stimuli create drastic changes in the classification where previous information does not help in class separation or understanding.

In this work, we created a sequential coarse-to-fine visual processing framework that is inspired by neuroscience findings on human perception and incorporates temporal dynamics in static image understanding. We also proposed a novel biology-inspired recurrent decomposing method, RSCD, to generate images in a CtF manner, and showed such processing helps the CtF framework be accurate and more robust even to adversarial attacks. In addition to RSCD, we evaluated the performance of our CtF framework on approximated decomposition, using JPEG compression and Gaussian smoothing. While these methods do not decompose over time and may have visually noticeable artifacts, they are fast and computationally efficient. Thus, they can be used in various applications with computational limitations.

\clearpage
\bibliographystyle{splncs04}
\bibliography{main}
\end{document}